\documentclass[10pt,twocolumn,letterpaper]{article}

\usepackage[pagenumbers]{cvpr} % To force page numbers, e.g. for an arXiv version

\usepackage{amsmath,amsfonts,bm}

\def\eqref#1{equation~\ref{#1}}
\def\1{\bm{1}}

\def\mK{{\bm{K}}}

\def\mQ{{\bm{Q}}}

\def\mS{{\bm{S}}}

\DeclareMathAlphabet{\mathsfit}{\encodingdefault}{\sfdefault}{m}{sl}
\SetMathAlphabet{\mathsfit}{bold}{\encodingdefault}{\sfdefault}{bx}{n}

\definecolor{mycyan}{RGB}{212, 239, 251}
\usepackage{colortbl}
\usepackage{xcolor}
\definecolor{mygray}{gray}{.9}
\definecolor{goldenrod}{RGB}{245,245,220}
\newlength\savewidth\newcommand\shline{\noalign{\global\savewidth\arrayrulewidth\global\arrayrulewidth 1pt}\hline\noalign{\global\arrayrulewidth\savewidth}}
\newcolumntype{a}{>{\columncolor{mygray}}c}
\usepackage{fontawesome5}
\usepackage{booktabs}
\usepackage{makecell}
\usepackage{fontawesome5}
\usepackage{lipsum}
\usepackage{comment}
\usepackage{multirow}
\usepackage{wrapfig}
\usepackage{algorithm}
\usepackage{algorithmic}
\usepackage{xfrac}  
\usepackage{adjustbox} % Add this line in the preamble

\usepackage{graphicx}
\usepackage{color}

\usepackage{amsthm}
\definecolor{darkgreen}{rgb}{0,0.7,0}

\definecolor{mygraytext}{gray}{.75}

\usepackage{tocloft}

\usepackage{etoolbox}
\makeatletter
\AfterEndEnvironment{algorithm}{\let\@algcomment\relax}
\AtEndEnvironment{algorithm}{\kern2pt\hrule\relax\vskip3pt\@algcomment}
\let\@algcomment\relax
\newcommand\algcomment[1]{\def\@algcomment{\footnotesize#1}}
\makeatother
\usepackage{listings}
\definecolor{cvprblue}{rgb}{0.21,0.49,0.74}
\usepackage[pagebackref,breaklinks,colorlinks,citecolor=cvprblue]{hyperref}

\title{VisionZip: Longer is Better but Not Necessary in Vision Language Models}

\author{
  Senqiao Yang$^{1}$ \hspace{0.35cm}
  Yukang Chen$^{1}$ \hspace{0.35cm}
  Zhuotao Tian$^{3}$\thanks{Corresponding to  tianzhuotao@gmail.com}\hspace{0.35cm}
  Chengyao Wang$^{1}$\hspace{0.35cm}
  Jingyao Li$^{1}$\hspace{0.35cm}
  Bei Yu$^{1}$\hspace{0.35cm}
  Jiaya Jia$^{1,2}$\\
  $^{1}$CUHK~~~
  $^{2}$HKUST~~~
  $^{3}$HITSZ~~~
}

\newcommand{\methodname}{VisionZip}

\newcommand{\mypara}[1]{\smallskip\noindent\textbf{#1}}
\begin{document}
\maketitle

\begin{abstract}
Recent advancements in vision-language models have enhanced performance by increasing the length of visual tokens, making them much longer than text tokens and significantly raising computational costs.
However, we observe that the visual tokens generated by popular vision encoders, such as CLIP and SigLIP, contain significant redundancy. 
To address this, we introduce~\methodname, a simple yet effective method that selects a set of informative tokens for input to the language model, reducing visual token redundancy and improving efficiency while maintaining model performance. 
The proposed VisionZip can be widely applied to image and video understanding tasks and is well-suited for multi-turn dialogues in real-world scenarios, where previous methods tend to underperform.
Experimental results show that \methodname~  outperforms the previous state-of-the-art method by at least 5\% performance gains across nearly all settings.
Moreover, our method significantly enhances model inference speed, improving the prefilling time by 8$\times$ and enabling the LLaVA-Next 13B model to infer faster than the LLaVA-Next 7B model while achieving better results.
Furthermore, we analyze the causes of this redundancy and encourage the community to focus on extracting better visual features rather than merely increasing token length. Our code is available at  \href{https://github.com/dvlab-research/VisionZip}{https://github.com/dvlab-research/VisionZip}.
\end{abstract}

\section{Introduction}
\label{sec:intro}

Recently, the advancement of Large Language Models~(LLMs)~\cite{touvron2023llama,achiam2023gpt,zhu2023minigpt,bai2023qwen} has led to significant progress in Vision Language Models~(VLMs)~\cite{li2024mini, li2023blip,liu2023improvedllava,chen2023sharegpt4v,Qwen-VL,longvila}. To integrate visual signals with textual semantics, existing VLMs typically utilize sequential visual representation, where images are converted into vision tokens and processed by an LLM decoder. Through modal alignment and instruction tuning, these VLMs adapt LLMs for the vision domain, leveraging their perception and reasoning capabilities.

However, the promising performance of VLMs largely relies on the large amount of visual tokens. For example, in LLaVA-1.5~\cite{liu2023improvedllava}, the number of visual tokens is 576, and in LLaVA-NeXT~\cite{liu2024llavanext}, a 672x672 image yield more than 576x5=2880 tokens, while the text tokens number only in the dozens to just over a hundred. 
These excessively long visual tokens consume a significant amount of memory and computation in the entire VLM, limiting the model's development in practical application scenarios such as edge computing, autonomous driving, and robotics~\cite{Yang_2024_CVPR,kim24openvla,yang2023lidar,qu2024mobile,yao2024minicpm,liu2024robomamba}.
Furthermore, based on many previous studies~\cite{kenton2019bert,dosovitskiy2020image,chen2023vlp,achiam2023gpt}, we know that the information contained in images is much sparser than in text. In contrast, the existing state-of-the-art VLMs have far more visual tokens than text tokens.
Hence, a natural question arises: \textbf{\textit{``Are all visual tokens necessary?"}}

\begin{figure}[t]
    \centering
    \includegraphics[width=1\linewidth]{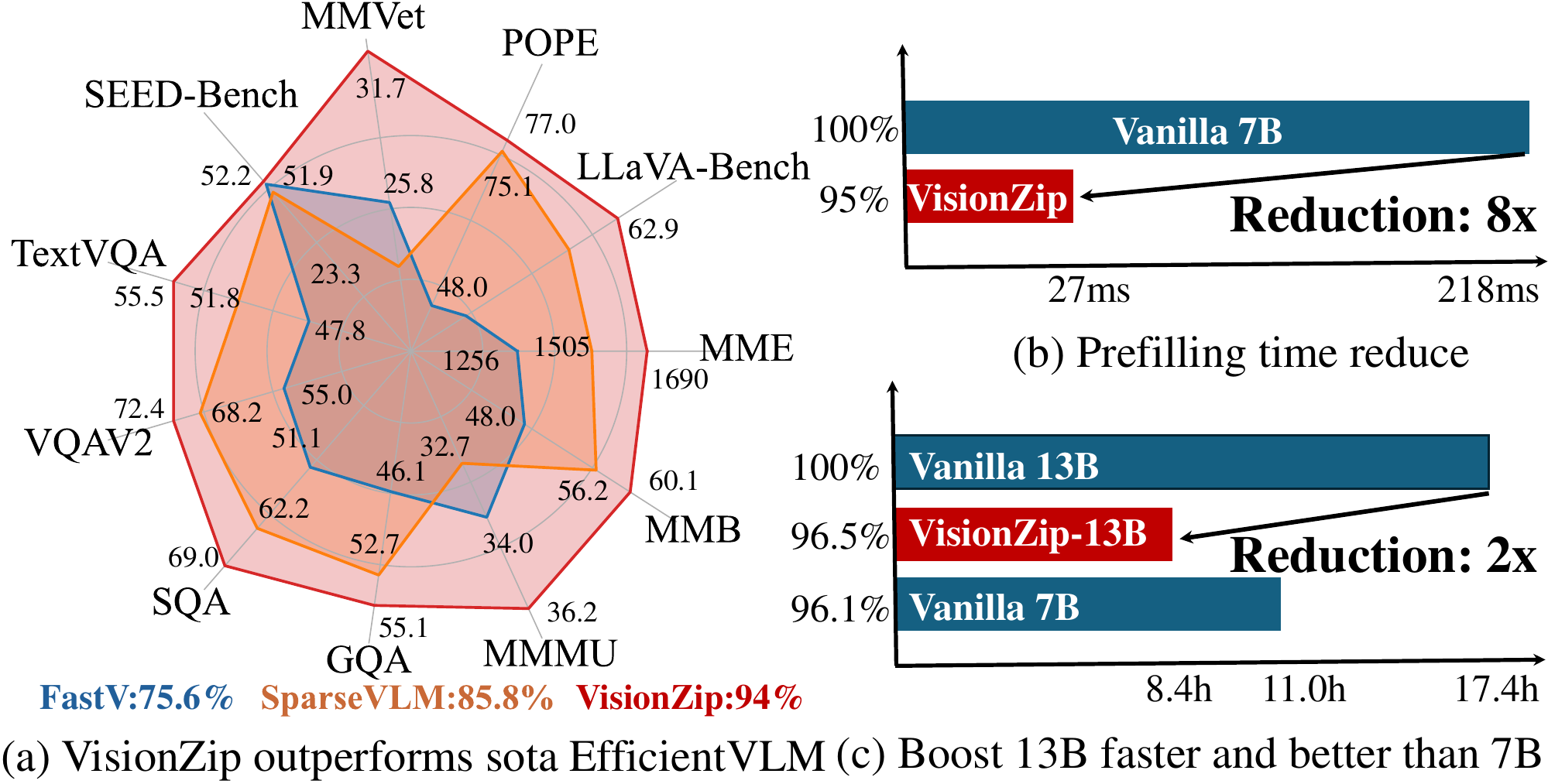}
    \caption{\textbf{VisionZip Performance and Efficiency.} (a) Our VisionZip significantly outperforms the current SOTA EfficientVLM model, like FastV, SparseVLM, achieving nearly 95\% of the performance with only 10\% of the tokens across 11 benchmarks on LLaVA-1.5. (b) VisionZip could reduce 8$\times$ prefilling time for LLaVA-NeXT 7B. (c) VisionZip reduces GPU inference time by 2$\times$ across 11 benchmarks, enabling the LLaVA-NeXT 13B model to infer faster than the 7B model while achieving better results.
    }
    % \vspace{-0.6cm}
    \label{fig:teaser-performance}
\end{figure}

\begin{figure}[t]
    \centering
    \includegraphics[width=1.0\linewidth]{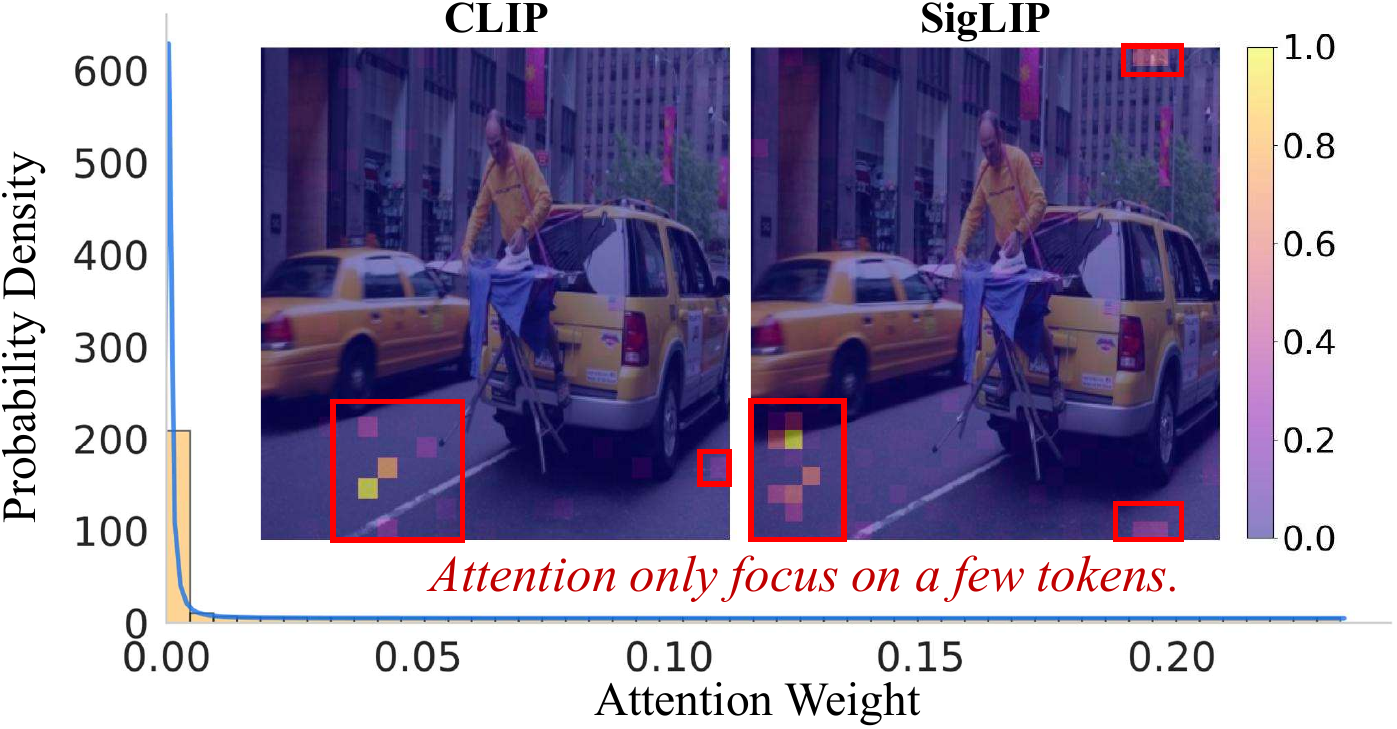}
 % \vspace{-0.6cm}
    \caption{\textbf{Redundancy Visualization.} The visualization and distribution statistics of attention scores show attention concentrated on only a few tokens, while many tokens display very low attention scores, indicating significant redundancy in the visual tokens.}
    \label{fig:teaser}
     % \vspace{-0.5cm}
\end{figure}

To explore this, we conduct a pilot study on the visual tokens generated by the widely used vision encoders, CLIP~\cite{radford2021learning} and SigLIP~\cite{zhai2023sigmoid}. As shown in Fig.~\ref{fig:teaser}, statistical and visual analysis reveal that only a few tokens receive high attention and contain a large amount of information, while most visual tokens receive minimal attention and aggregate limited information. 
Based on the observation, we can answer the question that
\textbf{\textit{there is a significant amount of redundancy in the visual tokens.}}
Details of this phenomenon's observation and the reasons behind it are provided in Sec.~\ref{sec:observe} and Sec.~\ref{sec:redundancy-reason}, respectively.

Based on this observation, we explore a solution to reduce visual token redundancy, aiming to improve efficiency without sacrificing performance. Specifically, we develop a text-agnostic method named \methodname~to extract more informative visual tokens for the LLM. Our method can be used in training-free, fine-tuning, or training from scratch.
Specifically, in training-free mode, we first select the dominant tokens, which receive significant attention and aggregate most of the image information. Then, to avoid missing small but potentially important details, we employ a token merging strategy, merging retained tokens based on their similarity to further extract informative contextual tokens.
In fine-tuning mode~(in Sec.~\ref{sec:readapt}), after selecting tokens to replace all raw visual tokens, the input token count decreases significantly, leading to a slight misalignment between the current visual input space and the LLM space. To enhance results and improve alignment, we fine-tune the projector layer for 30 minutes with minimal data, enabling the model to adapt to the reduced token count.

To demonstrate the effectiveness of our method, we apply the proposed \methodname~to popular VLM models and evaluate it on several benchmarks in Sec.~\ref{sec:exp}. As shown in Fig.~\ref{fig:teaser-performance}, the results indicate that even in a training-free scenario, our method significantly outperforms previous state-of-the-art methods in both speed and performance. Furthermore, \methodname~ can reduce pre-filling time by 8 times while retaining 95\% performance in LLaVA-NeXT 7B. The proposed \methodname~ also enables LLaVA-NeXT 13B to achieve better performance and faster inference than the LLaVA-NeXT 7B model. 
Finally, we analyze the causes of the redundancy and explain why the simple, text-agnostic \methodname~ achieves better performance than previous methods, highlighting its advantages in real-world deployment like multi-turn conversations in Sec.~\ref{sec:analysis}. 

\begin{figure*}[t]
    \centering
    \includegraphics[width=1.0\linewidth]{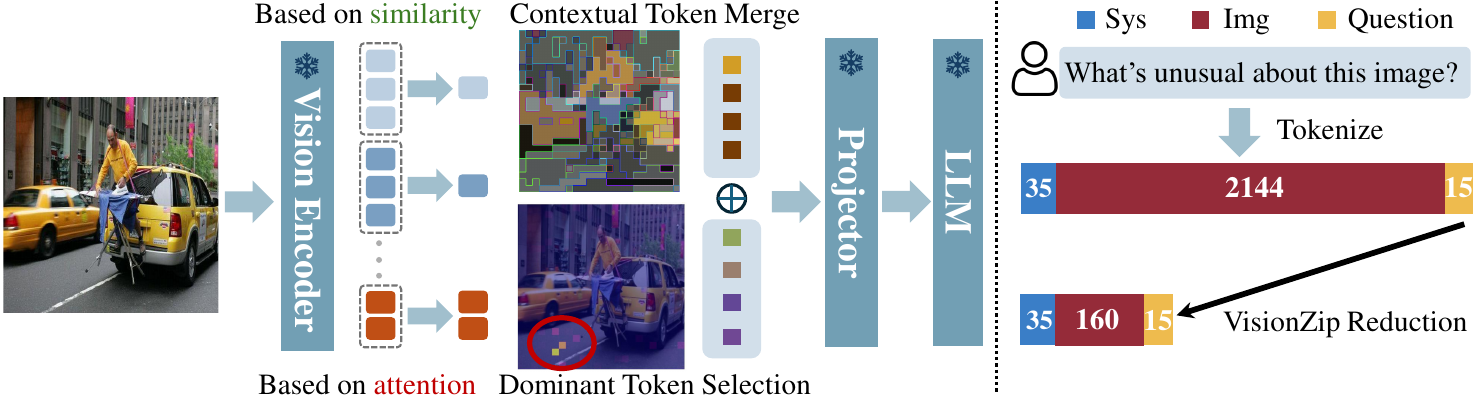}
     % \vspace{-0.2cm}
\caption{\textbf{Framework of \methodname.}  \methodname~ selects dominant tokens that aggregate substantial information based on visual token attention scores. Remaining tokens are merged based on semantic similarity to produce contextual tokens. ~\methodname\ is a training-free method significantly reduces the number of image tokens, accelerating inference while maintaining performance. With efficient fine-tuning of the projector, even better results can be achieved with minimal performance loss compared to using the full token.}
    \label{fig:framework}
     % \vspace{-0.4cm}
\end{figure*}

\section{\methodname}
\label{sec:visionzip}

In this section, we first explain the importance of reducing the number of visual tokens to improve model efficiency in Sec.~\ref{sec:pre}, and then present our observation of redundancy in Sec.~\ref{sec:observe}. After that, we detail the training-free method in Sec.~\ref{sec:IVZ}. Additionally, to help the model better adapt to variations in visual token length, we introduce Efficient Tuning in Sec.~\ref{sec:readapt}. Finally, we briefly discuss the widespread usage of \methodname. The overall architecture is shown in Fig.~\ref{fig:framework}.
\subsection{Preliminary}
\label{sec:pre}
\mypara{Architecture of VLM.}
The VLM architectures generally consist of three components: a visual encoder, a modality projector, and a LLM. The visual encoder, typically a pre-trained image encoder like CLIP's vision model, converts input images into visual tokens. The projector module aligns these visual tokens with the LLM's word embedding space, enabling the LLM to process visual data effectively. The LLM then integrates the aligned visual and textual information to generate responses.

\begin{algorithm}[t]
\caption{Pseudocode for Dominant Token Selection}
\label{algo:dominant}
\algcomment{\fontsize{7.2pt}{0em}\selectfont \texttt{cat}: concatenation;  \texttt{filter}: select the tokens based on the index.
%\vspace{-1.em}
}
\definecolor{codeblue}{rgb}{0.25,0.5,0.5}
\lstset{
  backgroundcolor=\color{white},
  basicstyle=\fontsize{7.2pt}{7.2pt}\ttfamily\selectfont,
  columns=fullflexible,
  breaklines=true,
  captionpos=b,
  commentstyle=\fontsize{7.2pt}{7.2pt}\color{codeblue},
  keywordstyle=\fontsize{7.2pt}{7.2pt},
%  frame=tb,
}
\begin{lstlisting}[language=python]
# B: batch size; S: sequence length 
# H: number of attention heads;
# K: number of target dominant tokens
# CLS_IDX: Index of the CLS token
# SELECT_LAYER: Selected layer for Visual Token

# set the output_attentions=True to get the attention
output = vision_tower(images, output_hidden_states=True, output_attentions=True)

#attn in shape (B, H, S, S)
attn = output.attentions[SELECT_LAYER]

#attn in shape (B, H, S, S)
vanilla_tokens = output.hidden_states[SELECT_LAYER]

#The attention received by each token
#If no CLS, use mean calculate received attention
attn_rec = attn[:, :, cls_idx, cls_idx+1:].sum(dim=1)

# Select K Dominant Tokens
_, topk_idx = attn_rec.topk(K, dim=1) 

# Concat with the CLS token
dominant_idx = cat(CLS_IDX, topk_idx+1)

# filter the Dominant Tokens
dominant_tokens = vanilla_tokens.filter(dominant_idx)

\end{lstlisting}

\end{algorithm}

\mypara{Computation Complexity.}
Evaluating the computational complexity of VLMs requires examining key components such as the self-attention mechanism and the feed-forward network (FFN). The total floating-point operations (FLOPs) can be expressed as: 
$$
\text{Total FLOPs} = T \times (4nd^2 + 2n^2d + 2ndm)
$$
where $T$ is the number of transformer layers, $n$ is the sequence length, $d$ is the hidden dimension size, and $m$ represents the intermediate size of the FFN. 
 
This equation shows that computational complexity is strongly influenced by the sequence length $n$. In typical VLM tasks, the sequence length is defined as $n = n_{\text{sys}} + n_{\text{img}} + n_{\text{question}}$, with $n_{\text{img}}$ often being much larger than the other two, sometimes by a factor of 20.  
Thus, \textbf{reducing $n_{\text{img}}$} is essential for improving the efficiency of VLMs. 
\subsection{Redundancy Observation}
\label{sec:observe}
In popular Vision Language Models like LLaVA and Mini-Gemini, the number of vision tokens far exceeds that of text tokens, consuming substantial computational resources. To assess whether all these tokens are necessary, we conducted a pilot study on the visual tokens generated by commonly used vision encoders, CLIP and SigLIP. 

Specifically, we randomly sampled one image and visualized the attention of each token from the Vision Encoder's -2 layer, which is the selected layer for obtaining input visual tokens in most VLMs, such as the LLaVA.
As shown in Fig.~\ref{fig:teaser}, both CLIP and SigLIP exhibit an attention pattern concentrated on a limited number of tokens, while the majority of visual tokens receive minimal attention. 
Furthermore, to demonstrate that the attention focusing on only a few tokens is a normal phenomenon, we analyze the distribution of attention weights on the TextVQA validation set. As shown in Fig.~\ref{fig:teaser},  most visual tokens receive very low attention, with weights close to zero, while only a few tokens hold higher attention weights.
To show this phenomenon’s prevalence, we include more visualizations in Appendix~\ref{sec:visual-redundancy}.

Based on this observation, we find that most visual tokens with low attention weights contribute little information and add significant redundancy. Only a few visual tokens aggregate a substantial amount of information and merit focused attention; we refer to these as the dominant visual tokens. 
Therefore, to reduce redundancy, we focus on selecting the most informative tokens—such as the dominant visual tokens—while discarding less informative ones to reduce the overall token count.

\begin{algorithm}[t]
\caption{Pseudocode for Contextual Tokens Merging.}
\label{algo:contexutual}
\algcomment{\fontsize{7.2pt}{0em}\selectfont 
%\vspace{-1.em}
\texttt{uniform\_split}: Uniformly sample the target tokens, and the rest are the merge tokens; \texttt{avg\_merge}: Average merge the tokens based on the assigned indices.
}
\definecolor{codeblue}{rgb}{0.25,0.5,0.5}
\lstset{
  backgroundcolor=\color{white},
  basicstyle=\fontsize{7.2pt}{7.2pt}\ttfamily\selectfont,
  columns=fullflexible,
  breaklines=true,
  captionpos=b,
  commentstyle=\fontsize{7.2pt}{7.2pt}\color{codeblue},
  keywordstyle=\fontsize{7.2pt}{7.2pt},
%  frame=tb,
}
\begin{lstlisting}[language=python]
# Remove dominant tokens
remaining = vanilla_tokens.mask(dominant_tokens)

# Split into target and merge tokens
# M represents the desired number of contextual tokens
targets, merge = uniform_split(remaining, M)

# Compute similarity based on the key values
simlarity = bmm(to_merge.K, targets.K.transpose(1, 2))

# Assign each merge token to the most similar target
assign_idx = simlarity.argmax(dim=2)

#  Merge by averaging
context_tokens = avg_merge(assign_idx, targets, merge)
\end{lstlisting}
\end{algorithm}

\subsection{Informative Visual Token Zip}
\label{sec:IVZ}

\mypara{Dominant Token Selection.}
To reduce redundancy by retaining only the most informative visual tokens and discarding less significant ones, the main challenge is identifying which tokens contribute most to the model’s performance.
We evaluate the importance of each visual token by examining its attention scores within the vision encoder. Specifically, we calculate the attention score as Eq.~\ref{eq:score}, 
\begin{equation}
 % \vspace{-0.15cm}
\label{eq:score}
\mS_h = \text{Softmax} \left( \frac{\mQ_h \mK_h^\top}{\sqrt{D_h}} \right),
 % \vspace{-0.1cm}
\end{equation}
where $\mS_h$ is the attention score of each head, $D_h$ is the head dimension, and $\mQ_h$ and $\mK_h$ represent query and key, respectively. Averaging across the head dimension, yields an aggregated attention matrix $\mS_{avg} \in \mathbb{R}^{B \times {SeqLen} \times {SeqLen}}$, reflecting  how each token attends to others.

For models with a CLS token, such as CLIP, which aggregates information from the entire image, we leverage the CLS token’s attention scores to identify key visual tokens. As shown in Algorithm~\ref{algo:dominant}, we select the tokens most attended to by the CLS token, as these typically contain the most relevant information.
For models without a CLS token, such as SigLIP, we calculate the average attention each token receives from all others in the sequence. Tokens with higher average attention are considered more significant and retained. We provide the details of it in Appendix~\ref{sec:no-cls}.

This process allows us to efficiently identify and retain the dominant visual tokens, as shown in Fig.~\ref{fig:framework}, these tokens contain almost all attention and aggregate substantial information of the total tokens.

\mypara{Contextual Tokens Merging.}
Although we have selected dominant tokens by evaluating their significance, and these dominant tokens contain most visual information, we merge the remaining tokens to avoid losing any small but potentially important information. 
Specifically, during self-attention calculation, the keys ($ K $) already summarize the information contained in each token. Therefore, as shown in Algorithm~\ref{algo:contexutual}, we first uniformly split the non-dominant tokens into target and merge tokens. We then use a similarity metric, such as the dot product, to identify the keys containing similar information. Finally, we merge the tokens that contain the most similar information, creating contextual tokens. As shown in Fig.~\ref{fig:framework}, these contextual tokens serve as highly informative tokens, containing the figure's semantic similarity information.

\subsection{Efficient Tuning}
\label{sec:readapt}
The Informative Visual Token Zip extracts highly informative tokens from the visual encoder and drops other tokens, thereby significantly reducing the token length input to the LLM, potentially by up to tenfold. However, this reduction in visual tokens can lead to a degree of misalignment, as the VLM model, originally trained on all full visual tokens, may struggle to adapt to the sudden decrease.

To bridge the gap between the visual and LLM spaces, we use minimal instruction tuning data to efficiently fine-tune the multimodal projector while keeping other components frozen, enhancing alignment between the vision and language spaces. Notably, the instruction tuning requires only 1/10 of the LLaVA-1.5 dataset and can be completed in just 30 minutes on 8 Nvidia A800 for LLaVA 1.5 7B. Notably, this process can also be implemented on 3090 GPUs, which is both resource-efficient and effective.

\subsection{Usage of VisionZip}
The ~\methodname~ can adapt to multiple tasks, not only for image and video understanding in Vision-Language Models but also for multi-turn conversations that previous efficient VLMs could not handle. Additionally, ~\methodname~ is easy to implement as it is text-agnostic, enabling compatibility with all existing LLM algorithms for acceleration. Furthermore, ~\methodname~ can be seen as a plug-and-play method for vision encoders, which preserves over 90\% of the original model's performance while saving 3 times runtime and memory. It can even allow a 13B VLM to achieve greater efficiency than a 7B VLM while maintaining superior performance.
We will show more details in Sec.~\ref{sec:advantage}.
\section{Experiments}
\label{sec:exp}
\renewcommand{\multirowsetup}{\centering}
\definecolor{mygray}{gray}{.92}
\definecolor{ForestGreen}{RGB}{34,139,34}
\newcommand{\fg}[1]{\mathbf{\mathcolor{ForestGreen}{#1}}}
\definecolor{Forestred}{RGB}{220,50,50}
\newcommand{\fr}[1]{\mathbf{\mathcolor{Forestred}{#1}}}
\begin{table*}[t]
    \centering
    % \hspace{2mm}
    
    \setlength{\tabcolsep}{2.8pt}
    \renewcommand{\arraystretch}{1.4}
    \footnotesize
	\centering
    \begin{tabular}{p{2.6cm}|c c c c c c c c c c c| c}
        \shline
        \textbf{Method} & \textbf{GQA} & \textbf{MMB} & \textbf{MME} & \textbf{POPE} & \textbf{SQA} & \textbf{VQA}$^{\text{V2}}$ & \textbf{VQA}$^{\text{Text}}$ & \textbf{MMMU}& \textbf{SEED} & \textbf{MMVet} & \textbf{LLaVA-B}  &\makecell[c]{\textbf{Avg}.}\\
        \shline
        \rowcolor{mygray}
        \multicolumn{13}{c}{\textit{Upper Bound, 576 Tokens} \ $\textbf{(100\%)}$}\\
        \multirow{2}*{Vanilla\texttt{\scriptsize{(CVPR24)}}} & 61.9 & 64.7 & 1862 & 85.9 & 69.5 & 78.5 & 58.2 & 36.3 & 58.6 & 31.1 & 66.8 & \multirow{2}*{100\%} \\
        ~ & 100\% & 100\% & 100\% & 100\% & 100\% & 100\% & 100\% & 100\% & 100\% & 100\% & 100\%  & ~ \\
        \hline

        \rowcolor{mygray}
        \multicolumn{13}{c}{\textit{Retain 192 Tokens} \ $\fg{(\downarrow 66.7\%)}$} \\

        \multirow{2}*{FastV \texttt{\scriptsize{(ECCV24)}}} & 52.7 & 61.2 & 1612 & 64.8 & 67.3 & 67.1 & 52.5 & 34.3 & 57.1 & 27.7 & 49.4 & \multirow{2}*{88.2\%} \\
        ~ & 85.1\% & 94.6\% & 86.6\% & 75.4\% & 96.8\% & 85.5\% & 90.2\% & 94.5\% &97.4\% &89.7\% &74.0\% & ~ \\
        \hline
        \multirow{2}*{SparseVLM\texttt{\scriptsize{(2024.10)}}} & 57.6 & 62.5 & 1721 & 83.6 & 69.1 & 75.6 & 56.1 & 33.8 & 55.8 & 31.5 & 66.1& \multirow{2}*{96.4\%} \\
        ~ & 93.1\% & 96.6\% & 92.4\% & 97.3\% & 99.4\% & 96.3\% & 96.4\% & 93.1\% & 95.2\% & 101.3\% & 99.0\%  & ~\\
        \hline
        \multirow{2}*{VisionZip} & 59.3 & 63.0 & 1782.6 & 85.3 & 68.9 & 76.8 &57.3 & 36.6 &56.4 &31.7& 67.7 & \multirow{2}*{\textbf{98.5\%}} \\
        ~ & 95.8\% & 97.4\% & 95.7\% & 99.3\% &99.1\% & 97.8\% & 98.5\% & 100.8\% & 96.2\% & 101.9\% & 101.3\% & ~\\
        \hline
        \multirow{2}*{VisionZip \ddag} & 60.1 & 63.4 & 1834 & 84.9 & 68.2 & 77.4 &57.8 & 36.2 & 57.1 & 32.6 & 66.7 & \multirow{2}*{$\fr{99.1\%}$} \\
        ~ & 97.1\% & 98.0\% & 98.5\% & 98.8\% &98.1\% & 98.6\% & 99.3\% & 99.7\% & 97.4\% &104.8\% &99.9\% &  ~\\
        \hline
        \rowcolor{mygray}
        \multicolumn{13}{c}{\textit{Retain 128 Tokens} \ $\fg{(\downarrow 77.8\%)}$}\\
        \multirow{2}*{FastV \texttt{\scriptsize{(ECCV24)}}} & 49.6 & 56.1 & 1490 & 59.6 & 60.2 & 61.8 & 50.6 & 34.9 & 55.9 & 28.1 & 52.0 &  \multirow{2}*{83.5\%}\\
        ~ & 80.1\% & 86.7\% & 80.0\% & 69.4\% & 86.6\% & 78.7\% & 86.9\% & 96.1\% & 95.4\% & 90.9\% & 77.8\%  & ~\\
        \hline
        \multirow{2}*{SparseVLM\texttt{\scriptsize{(2024.10)}}} & 56.0 & 60.0 & 1696 & 80.5 & 67.1 & 73.8 & 54.9 & 33.8 & 53.4 & 30& 62.7&   \multirow{2}*{93.4\%}\\
        ~ & 90.5\% & 92.7\% & 91.1\% & 93.7\% & 96.5\% & 94.0\% & 94.3\% & 93.1\% & 91.1\%& 96.5\%& 93.9\%& ~ \\
        \hline
        \multirow{2}*{VisionZip} & 57.6 & 62.0 & 1761.7 & 83.2 & 68.9 & 75.6 &56.8 & 37.9& 54.9& 32.6 &64.8& \multirow{2}*{\textbf{97.6\%}} \\
        ~ & 93.1\% & 95.8\% & 94.6\% & 96.9\% &99.1\% & 96.3\% & 97.6\% & 104.4\% &93.7\% & 104.8\% & 97.6\% ~ \\
        \hline
        \multirow{2}*{VisionZip \ddag} & 58.9 & 62.6 & 1823 & 83.7 & 68.3 & 76.6 &57.0  & 37.3 & 55.8 &  32.9 &  64.8 & \multirow{2}*{$\fr{98.4\%}$} \\
        ~ & 95.2\% & 96.8\% & 97.9\% & 97.4\% &98.3\% & 97.6\% & 97.9\% & 102.8\% & 95.2\% & 105.8\% & 97.0\% & ~\\
        \hline
        \rowcolor{mygray}
        \multicolumn{13}{c}{\textit{Retain 64 Tokens} \ $\fg{(\downarrow 88.9\%)}$}\\
        \multirow{2}*{FastV \texttt{\scriptsize{(ECCV24)}}} & 46.1 & 48.0 & 1256 & 48.0 & 51.1 & 55.0 & 47.8 & 34.0 & 51.9 & 25.8 & 46.1  & \multirow{2}*{75.6\%}\\
        ~ & 74.5\% & 74.2\% & 67.5\% & 55.9\% & 73.5\% & 70.1\% & 82.1\% & 93.7\% & 88.6\% &83.0\% & 69.0\% & ~\\
        \hline
        \multirow{2}*{SparseVLM\texttt{\scriptsize{(2024.10)}}} & 52.7 & 56.2 & 1505 & 75.1 & 62.2 & 68.2 & 51.8 & 32.7 & 51.1& 23.3& 57.5& \multirow{2}*{85.8\%} \\
        ~ & 85.1\% & 86.9\% & 80.8\% & 87.4\% & 89.4\% & 86.9\% & 89.0\% & 90.1\%& 87.2\% &74.5\% &86.1\% & \\
        \hline
        \multirow{2}*{VisionZip} & 55.1 & 60.1 & 1690 & 77.0 & 69.0 & 72.4 &55.5 & 36.2 & 52.2 & 31.7 &62.9& \multirow{2}*{\textbf{94.0\%}} \\
        ~ & 89.0\% & 92.9\% & 90.8\% & 89.6\% &99.3\% & 92.2\% & 95.4\% & 99.7\% & 89.1\% &101.9\% & 94.2\% & ~ \\
        \hline
        \multirow{2}*{VisionZip \ddag} & 57.0 & 61.5 & 1756 & 80.9& 68.8 & 74.2 &56.0 & 35.6 & 53.4 &30.2 &63.6 & \multirow{2}*{$\fr{95.2\%}$} \\
        ~ & 92.1\% & 95.1\% & 94.3\% & 94.2\% &99.0\% & 94.5\% & 96.2\% & 98.1\% & 91.1\% & 97.1\% &95.2\% \\
        \shline
	\end{tabular}
    % \vspace{-0.15cm}
	\caption{\textbf{Performance of ~\methodname\ on LLaVA 1.5.} The vanilla number of visual tokens is $576$. The first line of each method shows the raw benchmark accuracy, and the second line is the proportion relative to the upper limit. The last column is the average value. \methodname\ddag~ indicates that fine-tuning the multimodal projector with $1/10$ LLaVA-1.5 datasets, which takes 30 minutes for 8A800 GPU.}
	\label{tab:llava1-5}
    % \vspace{-0.5cm}
\end{table*}

\subsection{Effectiveness on Image Understanding}
\mypara{Evaluation Tasks.}
To show the effectiveness of our method on image understanding tasks, we conduct experiments on eleven widely used benchmarks~\cite{hudson2019gqa, liu2023mmbench,fu2023mme, li2023evaluating,lu2022learn, goyal2017making,singh2019towards,yue2023mmmu,li2023seed,yu2024mm,liu2023improvedllava} and compare our method with the existing sota methods, FastV~\cite{chen2024image} and SparseVLM~\cite{zhang2024sparsevlm}, which progressively reduce the number of visual tokens in the LLM forward process based on attention weights.
To further validate the generalizability of our method, we conduct experiments on various VLM with different architectures and resolutions. Due to space limitations, we present only a subset of results for LLaVA-1.5~\cite{liu2023improvedllava}, LLaVA-NeXT~\cite{liu2024llavanext}, and Mini-Gemini~\cite{li2024mini} in the main text and all results and implementation details can be found in Appendix~\ref{sec:sup-exp}.

\mypara{Results on LLaVA 1.5.}
As shown in Table~\ref{tab:llava1-5}, we deploy the proposed \methodname\ on LLaVA-1.5 and demonstrate its performance on image understanding tasks. \methodname\ represents our method being directly applied during the inference stage without additional training. \methodname\ddag\ denotes an efficient tuning for the cross-modality projector, requiring approximately 30 minutes on 8 A800 GPUs. This tuning can also be implemented on 3090 GPUs, making it both resource-efficient and effective.
To comprehensively assess performance, we present the results in percentage format for comparative analysis, with the vanilla model’s accuracy serving as the 100\% upper limit. Following the setup in~\cite{chen2024image,zhang2024sparsevlm}, we use three vision token count configurations ($192$, $128$, and $64$) to evaluate the advantages of our proposed \methodname.
When the visual tokens are reduced from $576$ to $192$, \methodname\ only decreases the average accuracy by $1.5\%$ without additional training, surpassing FastV \cite{chen2024image} by $10.3\%$ and SparseVLM~\cite{zhang2024sparsevlm} by $2.1\%$, respectively.
Furthermore, when only $64$ tokens remain, our method outperforms FastV \cite{chen2024image} and SparseVLM~\cite{zhang2024sparsevlm} by a significant margin of \textbf{18.4\%} and \textbf{8.2\%}, respectively.
Additionally, \methodname\ddag, which efficiently tunes the cross-modality projector, provides further performance improvements. As shown in Table~\ref{tab:llava1-5}, even with only 64 visual tokens retained, this efficient tuning boosts performance to \textbf{$95.2\%$}, representing only a $4.8\%$ decrease compared to the vanilla method using 10 times the visual tokens. 

An interesting phenomenon is that in certain benchmarks, such as MMVeT and MMMU, using \methodname to reduce the token count not only prevents performance degradation but also improves performance. We believe the reason is that the visual tokens are overly redundant, and this redundant information not only fails to improve model performance but may also act as noise, impacting the model's judgment and leading to performance degradation. We analyze this phenomenon in Sec.~\ref{sec:analysis}.
\renewcommand{\multirowsetup}{\centering}
\definecolor{mygray}{gray}{.92}
\definecolor{ForestGreen}{RGB}{34,139,34}
\definecolor{Forestred}{RGB}{220,50,50}

\begin{table}[t]
    \centering
    % Reduce column spacing
    \setlength{\tabcolsep}{1.5pt} % Default is approximately 6pt, reduced to 1.5pt
    % Increase font size
    \small % Increased from \footnotesize to \small
    \renewcommand{\arraystretch}{1.2} % Optional: slightly reduce row height
    %\begin{adjustbox}{max width=0.5\textwidth} 
    \resizebox{\columnwidth}{!}{
        \begin{tabular}{
            @{} 
            l | 
            *{7}{c} | 
            c
            @{}
        }
            \toprule
            \textbf{Method} & \textbf{GQA} & \textbf{MMB} & \textbf{MME} & \textbf{SQA} & \textbf{VQA}$^{\text{V2}}$ & \textbf{VQA}$^{\text{Text}}$ & \textbf{MMMU} & \makecell[c]{\textbf{Avg}.} \\
            \hline
            \rowcolor{mygray}
            \multicolumn{9}{c}{\textit{Upper Bound, 2880 Tokens} \ $\textbf{(100\%)}$} \\
            \multirow{2}{*}{Vanilla} & 64.2 & 67.9 & 1842 & 70.2 & 80.1 & 61.3 & 35.1 & \multirow{2}{*}{100\%} \\
            ~ & 100\% & 100\% & 100\% & 100\% & 100\% & 100\% & 100\% & ~ \\
            \hline

            \rowcolor{mygray}
            \multicolumn{9}{c}{\textit{Retain 640 Tokens} \ $\fg{(\downarrow 77.8\%)}$} \\
            \multirow{2}{*}{SparseVLM} & 60.3 & 65.7 & 1772 & 67.7 & 77.1 & 57.8 & 34.6 & \multirow{2}{*}{96.1\%} \\
            ~ & 93.9\% & 96.8\% & 96.2\% & 96.4\% & 96.3\% & 94.3\% & 98.6\% & ~ \\
            \hline
            \multirow{2}{*}{VisionZip} & 61.3 & 66.3 & 1787 & 68.1 & 79.1 & 60.2 & 34.7 & \multirow{2}{*}{\textbf{97.6\%}} \\
            ~ & 95.5\% & 97.6\% & 97.0\% & 97.0\% & 98.8\% & 98.2\% & 98.9\% & ~ \\
            \hline

            \multirow{2}{*}{VisionZip \ddag} & 62.4 & 65.9 & 1778 & 67.9 & 79.9 & 60.8 & 37.2 & \multirow{2}{*}{$\fr{98.9\%}$} \\
            ~ & 97.2\% & 97.1\% & 96.5\% & 96.7\% & 99.8\% & 99.2\% & 106.0\% & ~ \\
            \hline
            \rowcolor{mygray}
            \multicolumn{9}{c}{\textit{Retain 320 Tokens} \ $\fg{(\downarrow 88.9\%)}$} \\
            \multirow{2}{*}{SparseVLM} & 57.7 & 64.3 & 1694 & 67.3 & 73.4 & 55.9 & 34.4 & \multirow{2}{*}{93.3\%} \\
            ~ & 89.9\% & 94.7\% & 92.0\% & 95.9\% & 91.6\% & 91.2\% & 98.0\% & ~ \\
            \hline
            \multirow{2}{*}{VisionZip} & 59.3 & 63.1 & 1702 & 67.3 & 76.2 & 58.9 & 35.3 & \multirow{2}{*}{\textbf{95.0\%}} \\
            ~ & 92.3\% & 92.9\% & 92.4\% & 95.9\% & 95.1\% & 96.1\% & 100.5\% & ~ \\
            \hline
            \multirow{2}{*}{VisionZip \ddag} & 61.0 & 64.4 & 1770 & 67.5 & 78.4 & 59.3 & 38.0 & \multirow{2}{*}{$\fr{97.9\%}$} \\
            ~ & 95.0\% & 94.8\% & 96.1\% & 96.2\% & 97.9\% & 96.7\% & 108.3\% & ~ \\
            \hline
            \rowcolor{mygray}
            \multicolumn{9}{c}{\textit{Retain 160 Tokens} \ $\fg{(\downarrow 94.4\%)}$} \\
            \multirow{2}{*}{SparseVLM} & 51.2 & 63.1 & 1542 & 67.5 & 66.3 & 46.4 & 32.8 & \multirow{2}{*}{86.4\%} \\
            ~ & 79.8\% & 92.9\% & 83.7\% & 96.2\% & 82.8\% & 75.7\% & 93.4\% & ~ \\
            \hline
            \multirow{2}{*}{VisionZip} & 55.5 & 60.1 & 1630 & 68.3 & 71.4 & 56.2 & 36.1 & \multirow{2}{*}{\textbf{92.0\%}} \\
            ~ & 86.4\% & 88.5\% & 88.5\% & 97.3\% & 89.1\% & 91.7\% & 102.8\% & ~ \\
            \hline
            \multirow{2}{*}{VisionZip \ddag} & 58.2 & 63.9 & 1699 & 67.5 & 75.6 & 57.3 & 37.7 & \multirow{2}{*}{$\fr{95.5\%}$} \\
            ~ & 90.7\% & 94.1\% & 92.2\% & 96.2\% & 94.4\% & 93.5\% & 107.4\% & ~ \\
            \bottomrule
        \end{tabular}}
    %\end{adjustbox}
    % \vspace{-0.2cm}
    \caption{\textbf{Performance of VisionZip on LLaVA-NeXT.} The vanilla number of visual tokens is $2880$. For \methodname\ddag, we use $1/10$ LLaVA-1.5 datasets to fine-tune the multimodal projector.}
    \label{tab:llava1-6}
    % \vspace{-0.5cm}
\end{table}

\mypara{Results on LLaVA-NeXT. }
To further demonstrate the effectiveness of our proposed \methodname, we apply it to the more advanced, high-resolution-capable VLM, LLaVA-NeXT.
Compared to LLaVA 1.5, LLaVA-NeXT divides the image into four parts, resizes the original image, and converts it into five separate images. Each of these images is processed through the visual encoder to obtain visual tokens, which are then combined. While this approach further improves model performance, it significantly increases the number of visual tokens. Therefore, to enhance efficiency, we aim to use our method to reduce the number of visual tokens as much as possible without compromising model performance.
And we set the three vision token count configurations ($640$, $320$, and $160$) to evaluate the advantages of our proposed \methodname.
As shown in Table~\ref{tab:llava1-6}, our proposed VisionZip consistently maintains strong performance across three settings. Specifically, using only 640 tokens, our method achieves 97.6\% accuracy without any additional training cost. With minimal data used to tune the projector, VisionZip’s performance reaches 98.9\%, which is very close to that of the vanilla model.
Additionally, when the visual token count is reduced to only about 5\%, our method still achieves 92.0\% performance without any additional training and reaches 95.2\% after tuning, surpassing the previous state-of-the-art method, SparseVLM~\cite{zhang2024sparsevlm}, by 5.8\% and 9\%, respectively. And the full experiment results can be found in Appendix~\ref{sec:sup-exp}.

\begin{figure*}
    \centering
    \includegraphics[width=1.0\linewidth]{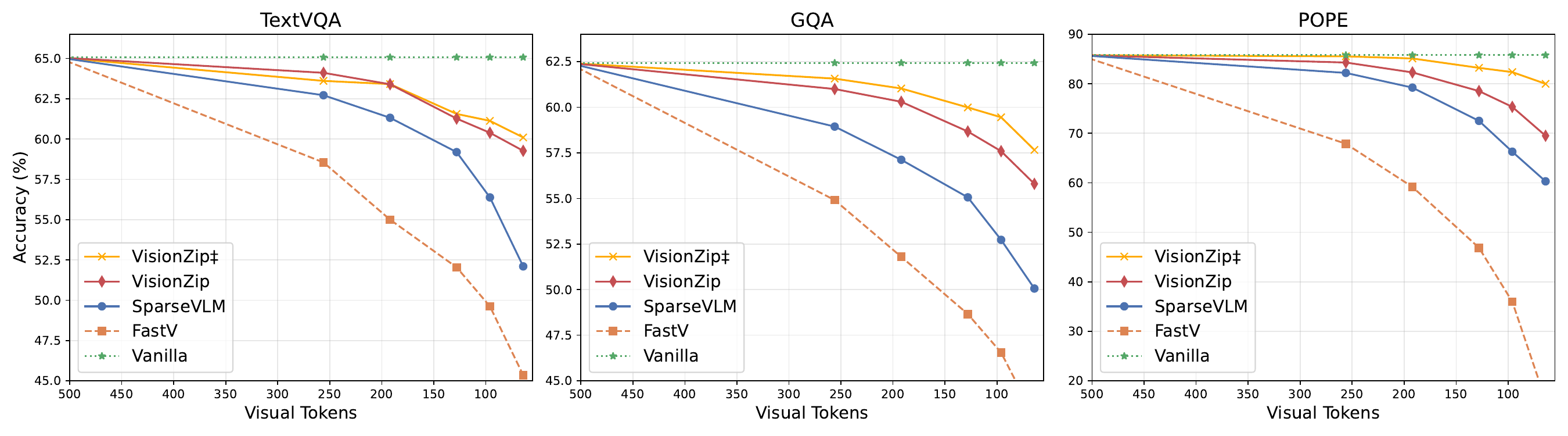}
    % \vspace{-0.9cm}
    \caption{Performance of VisionZip on the Mini-Gemini. }
    % \vspace{-0.6cm}
    \label{fig:mgm}
\end{figure*}

\mypara{Results on Mini-Gemini.}
We have verified the effectiveness of our method on the LLaVA Family VLMs, and we further validate our proposed VisionZip on Mini-Gemini, which introduces a LAION-pretrained ConvNeXt-L \cite{liu2022convnet} for high-resolution refinement, to demonstrate ~\methodname's effectiveness across different architectures.
As shown in Fig.~\ref{fig:mgm}, we visualize the performance change across different visual token counts on POPE, TextVQA, and GQA. It can be observed that as the number of tokens decreases, the gap between our method and the previous sota method increases sharply. These results further verify the effectiveness of our method across various model architectures and demonstrate the presence of visual token redundancy across multiple architectures. We discuss in Section \ref{sec:analysis} why our straightforward and easy-to-implement method VisionZip outperforms previous approaches.

\subsection{Effectiveness on Video Understanding}
\mypara{Evaluation Tasks.} 
We evaluate our method on four common video question-answering benchmarks: TGIF-QA \cite{jang2017tgif}, MSVD-QA \cite{xu2017video}, MSRVTT-QA \cite{xu2017video}, and ActivityNet-QA \cite{yu2019activityqa}, where video-question pairs exhibit significant length disparities. We follow the evaluation framework proposed by Video-LLaVA \cite{lin2023video}, utilizing ChatGPT score as key performance metrics. Further details are provided in Appendix~\ref{sec:sup-exp}.

\definecolor{mygray}{gray}{.90}

\renewcommand{\multirowsetup}{\centering}

\begin{table}[t]
    \centering
    \setlength{\tabcolsep}{1.5pt}
    \renewcommand{\arraystretch}{1.5}
    \small
    
    %\resizebox{\columnwidth}{!}{
        \begin{tabular}{c| c  c  c  c | c }
            \hline
            \multirow{1}{*}{\textbf{Method}} & 
            \multicolumn{1}{c}{\textbf{TGIF}} & 
            \multicolumn{1}{c}{\textbf{MSVD}} & 
            \multicolumn{1}{c}{\textbf{MSRVTT}} & 
            \multicolumn{1}{c|}{\textbf{ActivityNet}} & 
            \multicolumn{1}{c}{\textbf{Avg}} \\
            \cline{1-6}
            Video-LLaVA & 47.1 & 69.8 & 56.7 & 43.1 & 100.0\% \\
            \hline
            \multirow{2}{*}{FastV} & 
            23.1 & 38.0 & 19.3 & 30.6 & \multirow{2}{*}{52.1\%} \\
            ~ & 49.0\% & 54.4\% & 34.0\% & 71.0\% & ~ \\
            \hline
            \multirow{2}{*}{SparseVLM} & 
            44.7 & 68.2 & 31.0 & 42.6 & \multirow{2}{*}{86.5\%} \\
            ~ & 94.9\% & 97.7\% & 54.7\% & 98.8\% & ~ \\
            \hline
            \multirow{2}{*}{VisionZip} & 
            42.4 &  63.5 & 52.1 &43.0 & \multirow{2}{*}{\textbf{93.2\%}} \\
            ~ & 90.0\% & 91.0\% & 91.9\% & 99.8\% & ~ \\
            \hline
        \end{tabular}%}
        % \vspace{-0.2cm}
     \caption{\textbf{Performance of ~\methodname\ on Video-LLaVA.} The original Video-LLaVa's video token number is $2048$, while our VisionZip only retain the $136$ tokens.}
     \label{tab:videollava}
     % \vspace{-0.35cm}
\end{table}

\mypara{Results on Video-LLaVA. }
The vanilla Video-LLaVA~\cite{lin2023video} uses the Language-bind as vision encoder to encode 8 frames, with each frame containing 256 visual tokens, resulting in a total of 2048 visual tokens. Hence, we set the Video-LLaVA with 2048 video tokens as the upper bound, achieving an overall average accuracy of 100.0\% and a score of 0.00. 
To make a fair comparison, we follow the original settings for the baseline methods FastV~\cite{chen2024image}and SparseVLM~\cite{zhang2024sparsevlm}, pruning the visual tokens to 135. For each frame, we zip the visual tokens from 256 to 17, resulting in a total of 136 visual tokens for the entire video.
As shown in Table~\ref{tab:videollava}, our \methodname~ in training-free mode achieves 93.2\% accuracy across four benchmarks, outperforming the previous state-of-the-art method, SparseVLM, by 6.7\%. Moreover, on the largest benchmarks, MSRVTT, our method shows a significant improvement over SparseVLM by 37.2\%. Additionally, our method consistently exceeds 90\% performance across all benchmarks, further demonstrating \methodname's effectiveness and robustness. 

\subsection{Efficiency Analysis}
\renewcommand{\arraystretch}{1.02}
\begin{table}[t]
    \centering
    \setlength{\tabcolsep}{2pt} 
    \renewcommand{\arraystretch}{1.4} 
    \small

    % \vspace{2mm}
    \begin{adjustbox}{width=0.4\textwidth} 
        \begin{tabular}{l|c|c c|c c}
            \toprule
            \multirow{2}{*}{\textbf{Method}} & 
            \multirow{2}{*}{\textbf{Token}} & 
            \textbf{Total} & 
            \multirow{2}{*}{\textbf{\(\Delta\)}} & 
            \textbf{Prefilling}  & 
            \multirow{2}{*}{\textbf{\(\Delta\)}} \\
             & & 
             \textbf{Time$\downarrow$} &  & 
             \textbf{Time $\downarrow$} &  \\
            \midrule
            Baseline & 2880 & 2293s & - & 218ms & - \\
            \midrule
            FastV & 160 & 1792s & 1.3$\times$ & 119ms & 1.8$\times$  \\
            SparseVLM & 160 & {1895s} & 1.2$\times$ & 128ms & 1.7$\times$  \\
            \rowcolor{mygray}
            \methodname & 160 & \textbf{756s} & \textbf{3.0$\times$} & \textbf{27.8ms} & \textbf{7.8$\times$}  \\
            \bottomrule
        \end{tabular}
    \end{adjustbox}
    \caption{\textbf{Efficiency analysis of ~\methodname\ on LLaVA-NeXT 7B.} The detailed metrics include practical total time for one A800 GPU on POPE, Prefilling time(latency). \(\Delta\) denotes the reduction ratio.}
    \label{tab:efficiency}
    
    % \vspace{-0.5cm} 
\end{table}

Our proposed ~\methodname~ reduces the number of visual tokens input to the Large Language Model, resulting in significant efficiency and CUDA memory gains during inference. We conduct a comparative analysis of CUDA memory usage, and pre-filling time on LLaVA NeXT-7B, comparing our method with  FastV~\cite{chen2024image}, and SparseVLM~\cite{zhang2024sparsevlm}.

As shown in Table \ref{tab:efficiency}, we perform an inference efficiency analysis on a single NVIDIA A800-80GB, using POPE\cite{li2023evaluating} dataset a fair comparison. ``Prefilling time" refers to the latency required to generate the first token. The results show that our method not only surpasses previous approaches in performance but also maintains a substantial advantage over previous sota methods when reduced to the same number of tokens. On the POPE dataset, 
our method achieves a \textbf{3$\times$} improvement 
in overall time efficiency and a \textbf{7.8$\times$} improvement in prefilling time compared to the vanilla model.

\section{Analysis and Discussion}
\label{sec:analysis}
\begin{figure}
    \centering
    \includegraphics[width=1\linewidth]{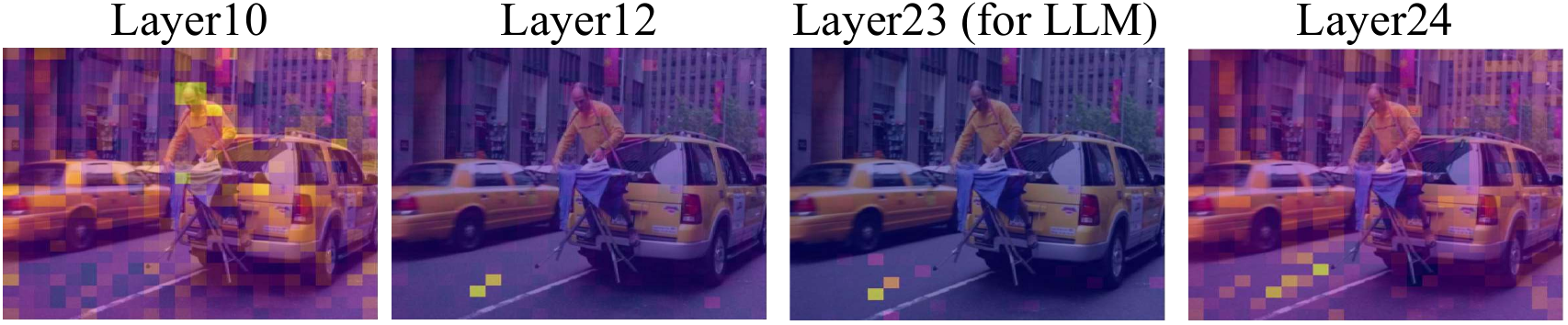}
    % \vspace{-0.6cm}
    \caption{Visualization of attention distribution across layers}
    % \vspace{-0.6cm}
    \label{fig:visualize-token}
\end{figure}
\subsection{Reasons of Redundancy in Visual Tokens}
\mypara{Visualization of the Redundancy. }
\label{sec:redundancy-reason}
Firstly, as shown in Fig.~\ref{fig:visualize-token}, we illustrate attention changes across layers. In early layers, attention is broadly distributed across the image, but by the middle layers, it suddenly converges onto a few tokens. With deeper layers, attention and information concentrate on a small set of dominant tokens, reaching peak concentration by the 23rd layer—used for visual token extraction for the LLM. Notably, attention is more dispersed in the final layer, as these tokens align with the CLIP text branch via contrastive loss, potentially limiting their representation of the original image. This is why VLM selects the second-to-last layer (-2 layer). Additional visualization results are in Appendix~\ref{sec:sup-visual}.

\mypara{Explanation. }
Current vision encoders are based on a transformer architecture that aggregates information between tokens through self-attention. We think that as the layer depth increases, instead of aggregating knowledge from all tokens, the model tends to ``shortcut" by concentrating information into a few proxy tokens. If a CLS token is present, the knowledge may further concentrate from these proxy tokens into the CLS token.
Moreover, using the function $ \text{softmax}(z_i) = \frac{e^{z_i}}{\sum_{j=1}^{n} e^{z_j}} $ to compute the model’s loss can intensify this effect. The derivative of this formula is as: 
\begin{equation}
    \frac{\partial \text{softmax}(z_i)}{\partial z_i} = \text{softmax}(z_i) \cdot (1 - \text{softmax}(z_i))  
\end{equation}

\begin{figure}
    \centering
    \includegraphics[width=1\linewidth]{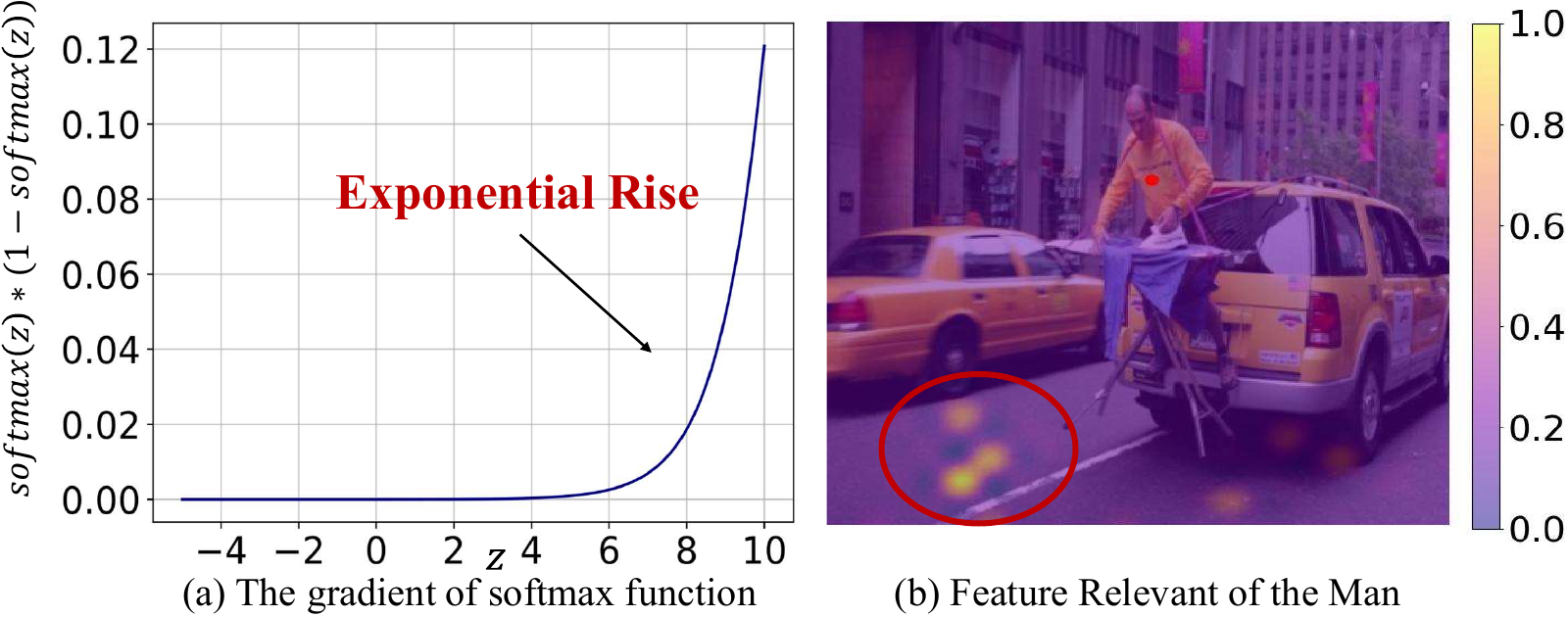}
    % \vspace{-0.5cm}
    \caption{Reason of redundancy and feature misalignment}
    \label{fig:softmax-misalign}
    % \vspace{-0.6cm}
\end{figure}

We illustrated this function in Fig.~\ref{fig:softmax-misalign}~(a), when $ z $ is large, the gradient becomes substantial in exponential rise, and when $ z $ is small, the gradient is almost negligible. This function makes regions of low attention even lower and high-attention areas even more prominent, ultimately concentrating information into a few tokens. ~\cite{xiao2023streamingllm} identified a similar phenomenon in LLM inference, naming it ``Attention Sink." ~\cite{shao2025explore} also observed a comparable effect in semantic segmentation, referring to it as the ``global token." 
\begin{table}[h!]
    \centering
    % \vspace{-0.4cm}
    \begin{tabular}{lccc}
        \toprule
        ~ & Token & Accuracy & $\Delta$ \\ 
        \midrule
        Baseline & 576$\rightarrow$64 & 51.1 & ~ \\ 
        $Ex1$ & 526$\rightarrow$64  & 46.4 & $-9.2\%$  \\ 
        $Ex2$ & 128$\rightarrow$64 & 52.5 & $+2.7\%$ \\ 
        \bottomrule
    \end{tabular}
    % \vspace{-0.2cm}
    \caption{Quantitative analysis for the feature misalignment}
    % \vspace{-0.6cm}
    \label{tab:example}
\end{table}

\subsection{Why ~\methodname~ Outperforms Previous Work?}
\textbf{Text-Relevant Efficient VLM.}
Existing sota methods for reducing visual redundancy to accelerate VLMs, such as FastV~\cite{chen2024image} and SparseVLM~\cite{zhang2024sparsevlm}, primarily rely on the LLM to identify text-relevant visual token. Specifically, they feed all visual tokens into the LLM and use attention between text and visual tokens across LLM layers for selection.

\mypara{Misalignment Due to the Pre-group Knowledge. }
While the text-relevant method appears promising, the visual tokens it selects often lack sufficient information. This limitation arises because the visual encoder aggregates visual information into a limited subset of high-attention tokens, leaving the remaining tokens with minimal informational content. As a result, tokens that should represent specific details are instead grouped into proxy tokens, losing their original incontext information. Furthermore, these proxy tokens tend to appear in peripheral or background areas rather than being positioned near the main subjects of the image. For instance, in Fig.~\ref{fig:softmax-misalign}~(b), the visual tokens most relevant to the person are not located on the person but are instead assigned to a proxy token situated on the road. This indicates that text-relevant methods often select tokens from elements like the man or the taxi, which actually contain significantly less informative content.

To further verify this, we performed two experiments on the TextVQA benchmark with SparseVLM, retaining 64 tokens, as shown in Table~\ref{tab:example}. In $Ex1$, we first masked 50 out of 576 total tokens, selecting the 50 tokens with the highest attention according to the vision encoder. From the remaining 526 tokens, SparseVLM was used to select the final set. This approach reduced performance from 51.1 to 46.4, a drop of approximately 9\%.
In $Ex2$, instead of providing all 576 tokens, we only supplied the top 128 tokens selected by ~\methodname~ to SparseVLM, which then filtered down to the final 64 tokens. This approach improved performance to 52.5, an increase of about 2.6\%. These results further verify that the text-relevant visual tokens are misaligned with the tokens where the Vision Encoder aggregates knowledge.

\subsection{The Advantage of the ~\methodname}
\label{sec:advantage}
\begin{figure}
    \centering
    \includegraphics[width=1\linewidth]{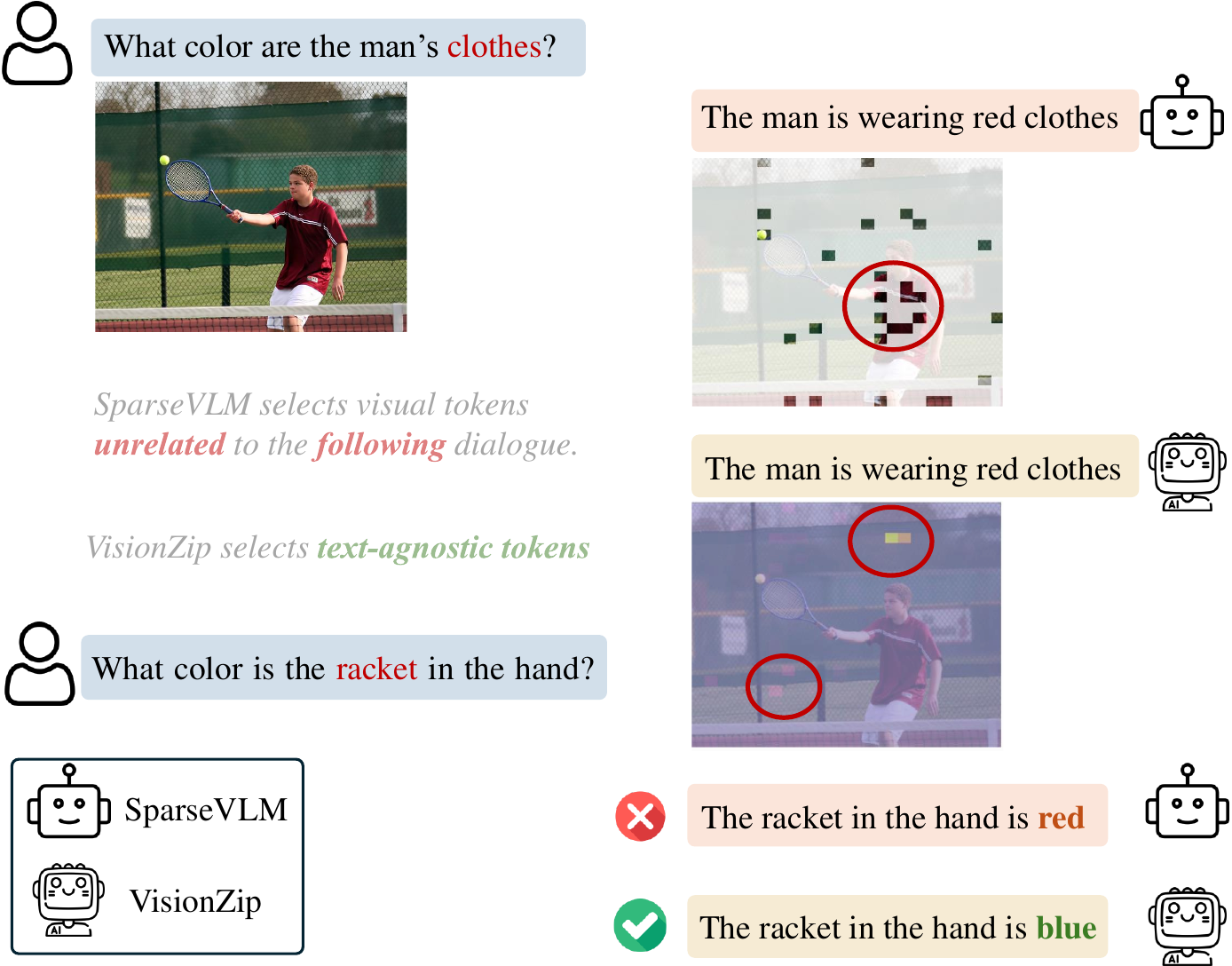}
    % \vspace{-0.4cm}
    \caption{Example comparison of VisionZip and previous text-relevant method in multi-turn conversation}
    \label{fig:case}
    % \vspace{-0.6cm}
\end{figure}

\textbf{Easy to deployment. }
Due to ~\methodname~ directly reducing the visual tokens before projecting them into the LLM, rather than gradually reducing them during the LLM forward process, it avoids extensive computation and memory consumption in the LLM’s shallow layers. As shown in Table~\ref{tab:advantage-quant}, our method is compatible with existing quantization techniques, maintaining performance while minimizing memory usage. Furthermore, our method enables the 13B model to be faster and perform better than the 7B model. As shown in Table~\ref{tab:advantage-acc}, our method significantly reduces the inference time of the 13B model, making it twice as fast as the vanilla 13B model and outperforming the vanilla 7B model in both performance and efficiency. Full results across 11 evaluation benchmarks are provided in Appendix~\ref{sec:sup-exp}. Additionally, ~\methodname~ is well-suited for integration with LLM acceleration optimization algorithms. 
\begin{table}[h!]
    \centering
    % \vspace{-0.2cm}
    \begin{minipage}{0.48\linewidth}
        \centering
        \renewcommand{\arraystretch}{1.2} % Adjust row height for better readability
        \begin{tabular}{p{1.55cm} p{1cm} p{0.4cm}} % Fixed column widths
            \toprule
            \textbf{Precision} & \textbf{Memory} & \textbf{Acc} \\ 
            \midrule
            7B-Full     & 18,952 & 70.2 \\ 
            13B-Full    & 36,721 & 73.5 \\ 
            13B-8bit-†  & 16,632 & 70.8 \\ 
            13B-4bit-†  & \textbf{10,176} & \textbf{70.3} \\ 
            \bottomrule
        \end{tabular}
        % \vspace{-0.2cm}
        \caption{Compatibility of VisionZip on various quantization levels for ScienceQA. † represents use of VisionZip.}
        \label{tab:advantage-quant}
    \end{minipage}%
    \hfill
    \begin{minipage}{0.48\linewidth}
        \centering
        \renewcommand{\arraystretch}{1.2} % Match row height for consistency
        \begin{tabular}{p{0.5cm} p{0.8cm} p{0.5cm}} % Fixed column widths
            \toprule
            \textbf{Size} & \textbf{Time} & \textbf{Acc} \\ 
            \midrule
            7B      & 1,714s & 61.3 \\ 
            13B     & 2,516s & 64.3 \\ 
            13B†    & \textbf{1,246s} & \textbf{62.2} \\ 
            \bottomrule
        \end{tabular}
        % \vspace{-0.2cm}
        \caption{VisionZip boosts the 13B model's performance and efficiency over the 7B model on TextVQA. † represents use of VisionZip.}
        \label{tab:advantage-acc}
    \end{minipage}
    % \vspace{-0.4cm}
\end{table}
\textbf{Advantage on multi-turn conversations.}
To better support real-world applications, current VLMs store the previous answer in the KV cache to enable multi-turn conversations, reducing the need to reprocess prior dialogue. However, as shown in Figure~\ref{fig:case}, prior text-relevant methods are unsuitable for multi-turn conversations. This is because the visual tokens selected and stored in the KV cache are closely related to the previous question but lack relevance to the current dialogue, leading to poor performance in multi-turn scenarios. In contrast, our ~\methodname\ selects the most informative visual tokens in a text-agnostic manner, making it more effective for multi-turn conversations. 

\section{Related Work}
\textbf{Vision-Language Models. }
Building on the success of large language models (LLMs) \cite{achiam2023gpt, touvron2023llama, bai2023qwen}, recent vision-language models (VLMs) \cite{liu2023improvedllava, chen2023internvl, li2024mini, tong2024cambrian} advance multimodal generation by processing extensive visual token sequences. Higher resolutions require exponentially more tokens; for example, LLaVA-NeXT processes $672 \times 672$ images into $2304$ tokens \cite{liu2023improvedllava}. Handling videos or multiple images increases token requirements, as seen in VideoLLaVA \cite{lin2023video} and Video-ChatGPT \cite{Maaz2023VideoChatGPT}. Hence, it’s essential to discuss more efficient ways to extract information from visual tokens, rather than merely increasing their length. The additional related work is shown in Appendix~\ref{sec:supp-related}.

\section{Conclusion}
In this paper, we analyze popular VLM models, noting that while increasing the length of visual tokens can improve performance, there is significant redundancy in current visual tokens. We propose a simple method,~\methodname, which reduces the number of visual tokens substantially while preserving model performance, thereby greatly enhancing computational efficiency. This method is broadly applicable to image and video understanding tasks and is suitable for multi-turn dialogue in practical applications.\methodname~ also suggests a future direction to develop vision encoders with lower redundancy capabilities to further improve VLM performance and handle longer video sequences.

{
    \small
    \bibliographystyle{ieeenat_fullname}
    \bibliography{main}
}
\newpage

% Before appendix, stop adding sections to the TOC (disable main body TOC)
\addtocontents{toc}{\protect\setcounter{tocdepth}{2}}

% Start the appendix
\appendix
\newpage

% Now allow sections in the appendix to appear in the TOC
\setcounter{tocdepth}{-1}  % Show sections and subsections in the appendix TOC
\hypersetup{linkcolor=cvprblue} 

\tableofcontents
\hypersetup{linkcolor=red}

\section{Further Discussion}
\subsection{Comparison with Text-relevant Efficient VLM}
We observe that most recent Efficient VLMs~\cite{chen2024image, zhang2024sparsevlm, xing2024pyramiddrop, he2024zipvl} utilize attention mechanisms between text tokens and visual tokens to determine which visual tokens should be retained, processing them during the LLM forward.
However, our method, \methodname, removes visual token redundancy before inputting them into the LLM. We will demonstrate our advantages from the following perspectives.

\mypara{Better Performance. } As shown in Table~\ref{tab:llava1-5},~\ref{tab:llava1-6},~\ref{tab:videollava} of the main paper, our~\methodname~achieves better performance in the training-free mode. This is because the Vision Encoder pre-groups the visual information into a few tokens, which often appear in the background or less prominent areas. However, when tokens are selected based on the semantic information of the text, the chosen tokens are often not the dominant tokens and carry less information, resulting in lower performance compared to~\methodname.
Additionally, to better demonstrate the misalignment caused by the Vision Encoder’s pre-grouping of information, we have created an interactive demo. As shown in Fig.~\ref{fig:gradio-demo}, the code for this demo will be published soon.

\mypara{More Efficient. } Our method reduces the redundancy of visual tokens before inputting them into the LLM, avoiding the heavy attention computation in the early layers of the LLM~(Sec.~\ref{sec:supp-efficiency}). 
Additionally, we observe that previous text-relevant Efficient VLMs require significant intermediate computations to determine which tokens need to be dropped during the LLM forward process. This leads to a noticeable increase in memory usage, sometimes exceeding that of the vanilla model. This issue is particularly evident in models like LLaVA-NeXT, where the number of visual tokens is substantial.

\mypara{More Application Scenarios. }
\methodname~operates outside the LLM, making it compatible with any existing LLM and applicable to all acceleration algorithms designed for LLMs. Furthermore, \methodname~is better suited for practical applications such as multi-turn conversations and other real-world scenarios.

\subsection{VisionZip for Non-CLS Vision Encoders}
\label{sec:no-cls}
Although most popular vision encoders, such as CLIP~\cite{radford2021learning}, OpenCLIP, and LanguageBind~\cite{zhu2023languagebind}, use the CLS token to aggregate information, a recently introduced vision encoder, SigLIP, does not include the CLS token.
To demonstrate the generalization of our proposed VisionZip, we explain how to apply it to Non-CLS Vision Encoders in this section.

Specifically, for the Dominant Token Selection,
we first calculate the attention score as shown in Eq.~\ref{eq:score1}, 
\begin{equation}
 % \vspace{-0.15cm}
\label{eq:score1}
\mS_h = \text{Softmax} \left( \frac{\mQ_h \mK_h^\top}{\sqrt{D_h}} \right),
 % \vspace{-0.1cm}
\end{equation}
where $\mS_h$ is the attention score of each head, and $D_h$ is the head dimension, $\mQ_h$ and $\mK_h$ represent query and key, respectively.
By averaging across the head dimension, we obtain an aggregated attention matrix $\mS_{avg} \in \mathbb{R}^{B \times {SeqLen} \times {SeqLen}}$, which reflects how each token attends to every other token.
The above process is similar to that of vision encoders with a CLS token, as described in the main text.

To identify key visual tokens, we calculate the average attention each token receives from all others in the sequence. Specifically, we compute the average along dim=1 of $\mS_{avg}$ to determine the degree to which each token is attended to by others, representing its importance.
Tokens with higher average attention are considered more significant and are retained.
We provide the pseudocode in Algorithm~\ref{algo:dominant-non}.

\begin{algorithm}[t]
\caption{Pseudocode for Dominant Token Selection-NO CLS Token}
\label{algo:dominant-non}
\algcomment{\fontsize{7.2pt}{0em}\selectfont \texttt{cat}: concatenation;  \texttt{filter}: select the tokens based on the index.
%\vspace{-1.em}
}
\definecolor{codeblue}{rgb}{0.25,0.5,0.5}
\lstset{
  backgroundcolor=\color{white},
  basicstyle=\fontsize{7.2pt}{7.2pt}\ttfamily\selectfont,
  columns=fullflexible,
  breaklines=true,
  captionpos=b,
  commentstyle=\fontsize{7.2pt}{7.2pt}\color{codeblue},
  keywordstyle=\fontsize{7.2pt}{7.2pt},
%  frame=tb,
}
\begin{lstlisting}[language=python]
# B: batch size; S: sequence length 
# H: number of attention heads;
# K: number of target dominant tokens
# CLS_IDX: Index of the CLS token
# SELECT_LAYER: Selected layer for Visual Token

# set the output_attentions=True to get the attention
output = vision_tower(images, output_hidden_states=True, output_attentions=True)

#attn in shape (B, H, S, S)
attn = output.attentions[SELECT_LAYER]

#attn in shape (B, H, S, S)
vanilla_tokens = output.hidden_states[SELECT_LAYER]

# no CLS token, use mean calculate received attention
attn_rec = attn.mean(dim=1).mean(dim=1) # (B, S)

# Select K Dominant Tokens
_, topk_idx = attn_rec.topk(K, dim=1) 

# filter the Dominant Tokens
dominant_tokens = vanilla_tokens.filter(topk_idx)

\end{lstlisting}

\end{algorithm}

\begin{table*}[h!]
\centering
\begin{tabular}{c|c|c|c|c|c|c}
\hline
 & \multicolumn{2}{c|}{\textbf{Retain 64}} & \multicolumn{2}{c|}{\textbf{Retain 128}} & \multicolumn{2}{c}{\textbf{Retain 192}} \\ \cline{2-7} 
 & \textbf{Dominant} & \textbf{Contextual} & \textbf{Dominant} & \textbf{Contextual} & \textbf{Dominant} & \textbf{Contextual} \\ \hline
\textbf{LLaVA-1.5} & 54 &10  & 108 & 20 & 162 & 30 \\ \hline
\textbf{Mini-Gemini} & 54 &10  & 108 & 20 & 162 & 30  \\ \hline

\end{tabular}
\caption{Token number settings for \methodname~ in LLaVA-1.5~\cite{liu2023improvedllava} and Mini-Gemini~\cite{li2024mini}}
\label{tab:token}
\end{table*}

\begin{table*}[h!]
\centering
\begin{tabular}{c|c|c|c|c|c|c}
\hline
 & \multicolumn{2}{c|}{\textbf{Retain 160}} & \multicolumn{2}{c|}{\textbf{Retain 320}} & \multicolumn{2}{c}{\textbf{Retain 640}} \\ \cline{2-7} 
 & \textbf{Dominant} & \textbf{Contextual} & \textbf{Dominant} & \textbf{Contextual} & \textbf{Dominant} & \textbf{Contextual} \\ \hline
\textbf{LLaVA NeXT} & 135 &25  & 270 & 50 & 540 & 100 \\ \hline

\end{tabular}
\caption{Token number settings for \methodname~ in LLaVA-NeXT~\cite{liu2024llavanext}}

\label{tab:token-next}
\end{table*}
\section{Additional Experiments}
\label{sec:sup-exp}
\subsection{Image Understanding}
\subsubsection{Implementation Details.}
\mypara{Environments.}
We conduct the inference on a single NVIDIA A800-80G GPU, while the fine-tuning process is performed on 8 NVIDIA A800-80G GPUs. Furthermore, to demonstrate the efficiency and effectiveness of our \methodname, the full training is conducted on 8 NVIDIA 3090-24G GPUs.

\mypara{Parameters.}
For the ~\methodname~ fine-tuning mode, we fine-tune only the cross-modality projector layer using a learning rate of $2e-5$, while keeping other components frozen.
For the ~\methodname~ training stage and inference mode, we follow the evaluation settings of the original model.

\mypara{Token Number.}
As shown in Table~\ref{tab:token}, for LLaVA-1.5 and Mini-Gemini, we present the number of dominant visual tokens and contextual visual tokens across three different configurations.
Additionally, for LLaVA-NeXT, which contains 5 subfigures, we provide the number of dominant visual tokens and contextual visual tokens across three different configurations in Table~\ref{tab:token-next}.

\renewcommand{\multirowsetup}{\centering}
\definecolor{mygray}{gray}{.92}
\definecolor{ForestGreen}{RGB}{34,139,34}
\definecolor{Forestred}{RGB}{220,50,50}
\begin{table*}[t]
    \centering
    % \hspace{2mm}
    
    \setlength{\tabcolsep}{2.8pt}
    \renewcommand{\arraystretch}{1.4}
    \footnotesize
	\centering
    \begin{tabular}{p{2.6cm}|c c c c c c c c c c c| c}
        \shline
        \textbf{Method} & \textbf{GQA} & \textbf{MMB} & \textbf{MME} & \textbf{POPE} & \textbf{SQA} & \textbf{VQA}$^{\text{V2}}$ & \textbf{VQA}$^{\text{Text}}$ & \textbf{MMMU}& \textbf{SEED-I} & \textbf{MMVet} & \textbf{LLaVA-B}  &\makecell[c]{\textbf{Avg}.}\\
        \shline
        \rowcolor{mygray}
        \multicolumn{13}{c}{\textit{Upper Bound, 576 Tokens} \ $\textbf{(100\%)}$}\\
        \multirow{2}*{Vanilla\texttt{\scriptsize{(CVPR24)}}} & 63.2 & 67.7 & 1818 & 85.9 & 72.8 & 80.0 & 61.3 & 36.4 & 66.9 & 35.3 & 70.8 & \multirow{2}*{100\%} \\
        ~ & 100\% & 100\% & 100\% & 100\% & 100\% & 100\% & 100\% & 100\% & 100\% & 100\% & 100\%  & ~ \\
        \hline

        \rowcolor{mygray}
        \multicolumn{13}{c}{\textit{Retain 192 Tokens} \ $\fg{(\downarrow 66.7\%)}$} \\

        \multirow{2}*{VisionZip} & 59.1 & 66.9 & 1754 & 85.1 & 73.5 & 78.1 &59.5 & 36.4 &65.2 &37.5& 77.5 & \multirow{2}*{\textbf{97.9\%}} \\
        ~ & 93.5\% & 98.8\% & 96.5\% & 99.1\% &101.0\% & 97.6\% & 97.1\% & 100\% & 97.5\% & 106.2\% & 109.5\% & ~\\
        \hline
        \multirow{2}*{VisionZip \ddag} & 61.6 & 67.1 & 1790 & 84.5 & 72.7 & 78.6 &59.9 & 36.4 & 66.1 & 37.7 & 73.9 & \multirow{2}*{$\fr{98.7\%}$} \\
        ~ & 97.5\% & 99.1\% & 98.5\% & 98.4\% &99.9\% & 98.3\% & 97.7\% & 100\% & 98.8\% &106.7\% &104.3\% &  ~\\
        \hline
        \rowcolor{mygray}
        \multicolumn{13}{c}{\textit{Retain 128 Tokens} \ $\fg{(\downarrow 77.8\%)}$}\\
        
        \multirow{2}*{VisionZip} & 57.9 & 66.7 & 1743 & 85.2 & 74.0 & 76.8 &58.7 & 36.1& 63.8& 37.5 &70.8& \multirow{2}*{\textbf{97.0\%}} \\
        ~ & 91.6\% & 98.5\% & 95.9\% & 99.2\% &101.6\% & 96.0\% & 95.8\% & 99.2\% &95.4\% & 106.2\% & 100\% ~ \\
        \hline
        \multirow{2}*{VisionZip \ddag} & 60.1 & 67.6 & 1736 & 83.8 & 73.0 & 77.6 &59.2  & 35.4 & 64.9 &  38.3 &  72.3 & \multirow{2}*{$\fr{97.4\%}$} \\
        ~ & 95.1\% & 99.9\% & 95.5\% & 97.6\% &100.2\% & 97.0\% & 96.6\% & 97.3\% & 97.0\% & 108.5\% & 102.1\% & ~\\
        \hline
        \rowcolor{mygray}
        \multicolumn{13}{c}{\textit{Retain 64 Tokens} \ $\fg{(\downarrow 88.9\%)}$}\\
        
        \multirow{2}*{VisionZip} & 56.2 & 64.9 & 1676 & 76.0 & 74.4 & 73.7 &57.4 & 36.4 & 60.4 & 33.9 &70.3& \multirow{2}*{\textbf{93.7\%}} \\
        ~ & 88.9\% & 95.9\% & 92.2\% & 88.5\% &102.2\% & 92.1\% & 93.3\% & 100\% & 90.3\% &96.0\% & 99.3\% & ~ \\
        \hline
        \multirow{2}*{VisionZip \ddag} & 58.1 & 65.6 & 1671 & 81.6& 72.3 & 75.2 &58.5 & 35.3 & 61.4 &36.7 &68.7 & \multirow{2}*{$\fr{94.8\%}$} \\
        ~ & 91.9\% & 96.9\% & 91.9\% & 95.0\% &99.3\% & 94.0\% & 95.4\% & 97.0\% & 91.8\% & 104.0\% &97.0\% \\

        \shline
	\end{tabular}
    % \vspace{-0.15cm}
	\caption{\textbf{Performance of ~\methodname\ on LLaVA 1.5 13B.} The vanilla number of visual tokens is $576$. The first line of each method shows the raw benchmark accuracy, and the second line is the proportion relative to the upper limit. The last column is the average value. \methodname\ddag~indicates that fine-tuning the multimodal projector with $1/10$ LLaVA-1.5 datasets. SEED-I represents SEED-IMG, which uses the metric from LMMs-Eval~\cite{zhang2024lmms}. The Avg calculation process does not include the results from LLaVA-B and MMVet, as the benchmark is small and the results are not stable.}
	\label{tab:llava1-5-13b}
    % \vspace{-0.5cm}
\end{table*}

\renewcommand{\multirowsetup}{\centering}
\definecolor{mygray}{gray}{.92}
\definecolor{ForestGreen}{RGB}{34,139,34}

\definecolor{Forestred}{RGB}{220,50,50}

\begin{table*}[t]
    \centering
    % \hspace{2mm}
    
    \setlength{\tabcolsep}{2.8pt}
    \renewcommand{\arraystretch}{1.4}
    \footnotesize
	\centering
    \begin{tabular}{p{2.6cm}|c c c c c c c c c c c c| c}
        \shline
        \textbf{Method} & \textbf{GQA} & \textbf{MMB} & \textbf{MME} & \textbf{POPE} & \textbf{SQA} & \textbf{VQA}$^{\text{V2}}$ & \textbf{VQA}$^{\text{Text}}$ & \textbf{MMMU}& \textbf{SEED} & \textbf{MMVet} & \textbf{VizWiz} & \textbf{LLaVA-B}  &\makecell[c]{\textbf{Avg}.}\\
        \shline
        \multirow{2}*{Vanilla\texttt{\scriptsize{(CVPR24)}}} & 61.9 & 64.7 & 1862 & 85.9 & 69.5& 78.5 & 58.2 & 36.3 & 58.6 & 31.1 & 50.0 &66.8 & \multirow{2}*{100\%} \\
        ~ & 100\% & 100\% & 100\% & 100\% & 100\% & 100\% & 100\% & 100\% & 100\% & 100\% & 100\% & 100\%  & ~ \\
        \hline

        \multirow{2}*{~\methodname\ 192 Tokens} & 61.5 & 67.4 & 1820 & 85.2 & 69.3 & 78.5&57.8 & 36.1 &59.6 &33& 52.6 & 71.3 & \multirow{2}*{\textbf{$\fr{100.6\%}$}} \\
        ~ & 99.4\% & 104.2\% & 97.7\% & 99.2\% &99.7\% & 100\% & 99.3\% & 99.4\% & 101.7\% & 106.1\% &105.2& 106.7\% & ~\\

        \hline

        \multirow{2}*{~\methodname\ 128 Tokens} & 60.0& 66.6 & 1814 & 84.3 & 69.4 & 77.8&57.6 & 36.9 &59.0&31.4& 49.9 & 66.7 & \multirow{2}*{\textbf{{99.6\%}}} \\
        ~ & 96.9\% & 102.9\% & 97.4\% & 98.1\% &99.9\% & 99.1\% & 99.0\% & 101.7\% & 100.7\% & 101\% & 99.8\%& 99.9\% & ~\\

        \hline

        \multirow{2}*{~\methodname\ 64 Tokens} & 58.9 & 63.7 & 1785 & 84.1 & 69.3 & 76.0&57.1 & 36.2 &55.8 & 29.9 & 46.8& 63.5 & \multirow{2}*{\textbf{97.1\%}} \\
        ~ & 95.2\% & 98.5\% & 95.9\% & 97.9\% &99.7\% & 96.8\% & 98.1\% & 99.7\% & 95.2\% & 96.1\% & 93.6\% &95.1\% & ~\\

        \shline
	\end{tabular}
    % \vspace{-0.15cm}
	\caption{\textbf{Using ~\methodname\ train the LLaVA 1.5 7B.} The vanilla number of visual tokens is $576$. The first line of each method shows the raw benchmark accuracy, and the second line is the proportion relative to the upper limit. The last column is the average value. \methodname\ddag~ indicates that fine-tuning the multimodal projector with $1/10$ LLaVA-1.5 datasets. The Avg calculation process does not include the results from LLaVA-B and MMVet, as the benchmark is small and the results are not stable.}
	\label{tab:llava1-5-train-supp}
    % \vspace{-0.5cm}
\end{table*}

\renewcommand{\multirowsetup}{\centering}
\definecolor{mygray}{gray}{.92}
\definecolor{ForestGreen}{RGB}{34,139,34}
\definecolor{Forestred}{RGB}{220,50,50}
\begin{table*}[t]
    \centering
    % \hspace{2mm}
    
    \setlength{\tabcolsep}{2.8pt}
    \renewcommand{\arraystretch}{1.4}
    \footnotesize
	\centering
    \begin{tabular}{p{2.6cm}|c c c c c c c c c| c}
        \shline
        \textbf{Method} & \textbf{GQA} & \textbf{MMB} & \textbf{MME} & \textbf{POPE} & \textbf{SQA} & \textbf{VQA}$^{\text{V2}}$ & \textbf{VQA}$^{\text{Text}}$ & \textbf{MMMU}& \textbf{SEED-I}   &\makecell[c]{\textbf{Avg}.}\\
        \shline
        \rowcolor{mygray}
        \multicolumn{11}{c}{\textit{Upper Bound, 2880 Tokens} \ $\textbf{(100\%)}$}\\
        \multirow{2}*{Vanilla} & 64.2 & 67.9 & 1842 &  86.4  & 70.2 & 80.1 & 61.3 & 35.1 & 70.2 & \multirow{2}*{100\%} \\
        ~ & 100\% & 100\% & 100\% & 100\% & 100\% & 100\% & 100\% & 100\% & 100\%   & ~ \\
        \hline

        \rowcolor{mygray}
        \multicolumn{11}{c}{\textit{Retain 640 Tokens} \ $\fg{(\downarrow 77.8\%)}$} \\

        \multirow{2}*{VisionZip} & 61.3 & 66.3 & 1787 & 86.3 & 68.1& 79.1 &60.2 & 34.7  &66.7  & \multirow{2}*{\textbf{97.5\%}} \\
        ~ & 95.5\% & 97.6\% & 97.0\% & 99.9\% &97.0\% & 98.8\% & 98.2\% & 98.9\% & 95.0\%& ~\\
        \hline
        \multirow{2}*{VisionZip \ddag} &  62.4 & 65.9 & 1778& 87.6  &67.9& 79.9 &60.8 & 37.2 & 67.8 & \multirow{2}*{$\fr{98.9\%}$} \\
        ~ & 97.2\% & 97.1\% & 96.5\% & 101.4\% &96.7\% & 99.8\% & 99.2\% & 106.0\% & 96.6\%& ~\\
        \hline
        \rowcolor{mygray}
        \multicolumn{11}{c}{\textit{Retain 320 Tokens} \ $\fg{(\downarrow 88.9\%)}$}\\
        
        \multirow{2}*{VisionZip} & 59.3 & 63.1 & 1702 & 82.1 &67.3& 76.2 &58.9 & 35.3& 63.4& \multirow{2}*{\textbf{94.5\%}} \\
        ~ & 92.3\% & 92.9\% & 92.4\% & 95.0\% & 95.9\% & 95.1\% & 96.1\% & 100.5\% & 90.3\% & ~\\
        \hline
        \multirow{2}*{VisionZip \ddag} & 61.0 & 64.4 & 1770 & 86.2 &67.5& 78.4 &59.3 & 38.0 & 65.9 & \multirow{2}*{$\fr{97.6\%}$} \\
        ~ & 95.0\% & 94.8\% & 96.1\% & 99.8\% & 96.2\% & 97.9\% & 96.7\% & 108.3\% & 93.9\% & ~ \\
        \hline
        \rowcolor{mygray}
        \multicolumn{11}{c}{\textit{Retain 160 Tokens} \ $\fg{(\downarrow 94.4\%)}$}\\
        
        \multirow{2}*{VisionZip} & 55.5 & 60.1 & 1630 & 74.8 &68.3& 71.4 &56.2 & 36.1 & 58.3 & \multirow{2}*{\textbf{91.5\%}} \\
        %     GQA  MMB   MME    POPE  SQA   VQAV2 VQAText  MMMU
        %4bit      60.1 1680.4        68.0  72.1    55.2    
        %8bit 55.1 59.7 1713.16 80.0  68.96 72.4    55.53   35.22
        ~ & 86.4\% & 88.5\% & 88.5\% & 86.6\% &97.3\% & 89.1\% & 91.7\% & 102.8\% & 83.0\% &~ \\
        \hline
        \multirow{2}*{VisionZip \ddag} & 58.2 & 63.9 & 1699 & 83.4 &67.5& 75.6 &57.3 & 37.7 & 62.9  & \multirow{2}*{$\fr{95.0\%}$} \\
            ~ & 90.7\% & 94.1\% & 92.2\% &  96.5\% & 96.2\% & 94.4\% & 93.5\% & 107.4\% & 89.6\% &~ \\
        % \multirow{2}*{VisionZip (tuning)} & 57.0 & 61.5 & 1779 & 83.0 & 68.8 & 74.2 &56.0 & 35.6 & \multirow{2}*{$\fr{95.7\%}$} \\

        \shline
	\end{tabular}
    % \vspace{-0.15cm}
	\caption{\textbf{Performance of ~\methodname\ on LLaVA NeXT 7B.} The vanilla number of visual tokens is $2880$. The first line of each method shows the raw benchmark accuracy, and the second line is the proportion relative to the upper limit. The last column is the average value. \methodname\ddag~indicates that fine-tuning the multimodal projector with $1/10$ LLaVA-1.5 datasets. SEED-I represents SEED-IMG, which uses the metric from LMMs-Eval~\cite{zhang2024lmms}.}
	\label{tab:llava1-6-7b-full}
    % \vspace{-0.5cm}
\end{table*}

\renewcommand{\multirowsetup}{\centering}
\definecolor{mygray}{gray}{.92}
\definecolor{ForestGreen}{RGB}{34,139,34}
\definecolor{Forestred}{RGB}{220,50,50}
\begin{table*}[t]
    \centering
    % \hspace{2mm}
    
    \setlength{\tabcolsep}{2.8pt}
    \renewcommand{\arraystretch}{1.4}
    \footnotesize
	\centering
    \begin{tabular}{p{2.6cm}|c c c c c c c c c| c}
        \shline
        \textbf{Method} & \textbf{GQA} & \textbf{MMB} & \textbf{MME} & \textbf{POPE} & \textbf{SQA} & \textbf{VQA}$^{\text{V2}}$ & \textbf{VQA}$^{\text{Text}}$ & \textbf{MMMU}& \textbf{SEED-I}   &\makecell[c]{\textbf{Avg}.}\\
        \shline
        \rowcolor{mygray}
        \multicolumn{11}{c}{\textit{Upper Bound, 2880 Tokens} \ $\textbf{(100\%)}$}\\
        \multirow{2}*{Vanilla 13B} &65.4& 70.0 & 1901 &  86.2  & 73.5 & 81.8 & 64.3 & 36.2 & 71.9 & \multirow{2}*{100\%} \\
        ~ & 100\% & 100\% & 100\% & 100\% & 100\% & 100\% & 100\% & 100\% & 100\%   & ~ \\
        \hline
        \multirow{2}*{Vanilla 7B} & 64.2 & 67.9 & 1842 &  86.4  & 70.2 & 80.1 & 61.3 & 35.1 & 70.2 & \multirow{2}*{97.2\%} \\
        ~ & 98.2\% & 96.3\% & 96.9\% & 100.2\% & 95.5\% & 97.9\% & 95.3\% & 97.0\% & 97.6\%   & ~ \\
        \hline

        \rowcolor{mygray}
        \multicolumn{11}{c}{\textit{Retain 640 Tokens} \ $\fg{(\downarrow 77.8\%)}$} \\

        \multirow{2}*{VisionZip} & 63.0 & 68.6 & 1871 &85.7 & 71.2& 79.7 &62.2 & 36.4  &68.8 & \multirow{2}*{\textbf{97.5\%}} \\
        ~ & 96.3\% & 98.0\% & 98.4\% &99.4\% &96.7\% & 96.9\% & 96.7\% & 100.5\% & 95.7\%& ~\\
        \hline
        \multirow{2}*{VisionZip \ddag} &  63.7& 66.6 & 1829& 86.3  &73.2& 81.2 &64.4 & 38.1 & 69.2 & \multirow{2}*{$\fr{98.8\%}$} \\
        ~ & 97.4\% & 95.1\% & 96.2\% &100.1\% &99.6\% & 99.3\% & 100.2\% & 105.2\% & 96.2\%& ~\\
        \hline
        \rowcolor{mygray}
        \multicolumn{11}{c}{\textit{Retain 320 Tokens} \ $\fg{(\downarrow 88.9\%)}$}\\
        
        \multirow{2}*{VisionZip} & 60.7 & 67.2 &1805 & 82.0 &70.3& 76.8 &60.9 & 35.6& 65.2& \multirow{2}*{\textbf{94.7\%}} \\
        ~ & 92.8\% & 96.0\% & 95.0\% &95.1\% & 95.6\% & 93.9\% & 94.7\% & 98.3\% & 90.7\% & ~\\
        \hline
        \multirow{2}*{VisionZip \ddag} & 62.5 &66.9 & 1861  &85.7& 72.7 &80.0 & 63.2 & 36.9 & 67.9& \multirow{2}*{$\fr{97.8\%}$} \\
        ~ & 95.6\% & 95.6\% & 97.9\% & 99.4\% & 98.9\% & 97.8\% & 98.3\% & 101.9\% & 94.4\% & ~ \\
        \hline
        \rowcolor{mygray}
        \multicolumn{11}{c}{\textit{Retain 160 Tokens} \ $\fg{(\downarrow 94.4\%)}$}\\
        
        \multirow{2}*{VisionZip} & 57.8 & 64.9 & 1739 & 76.6 &69.3& 72.4 &58.4 & 37.0 & 61.1& \multirow{2}*{\textbf{91.3\%}} \\
        ~ & 88.4\% & 92.7\% & 91.5\% & 88.9\% & 94.3\% & 88.5\% & 90.8\% & 102.2\% & 84.8\% & ~ \\
        \hline
        \multirow{2}*{VisionZip \ddag} & 59.7 &65.3 & 1766 & 84.0 &72.0& 77.6 &60.8 & 36.0 & 64.4 & \multirow{2}*{$\fr{94.6\%}$} \\
        ~ & 91.3\% & 93.3\% & 92.9\% & 97.4\% & 98.0\% & 94.9\% & 94.6\% & 99.4\% & 89.6\% & ~ \\
        \shline
	\end{tabular}
    % \vspace{-0.15cm}
	\caption{\textbf{Performance of ~\methodname\ on LLaVA NeXT 13B.} The vanilla number of visual tokens is $2880$. The first line of each method shows the raw benchmark accuracy, and the second line is the proportion relative to the upper limit. The last column is the average value. \methodname\ddag~indicates that fine-tuning the multimodal projector with $1/10$ LLaVA-1.5 datasets. SEED-I represents SEED-IMG, which uses the metric from LMMs-Eval~\cite{zhang2024lmms}.}
	\label{tab:llava1-6-13b-full}
    % \vspace{-0.5cm}
\end{table*}

\renewcommand{\multirowsetup}{\centering}
\definecolor{mygray}{gray}{.92}
\definecolor{ForestGreen}{RGB}{34,139,34}
\definecolor{Forestred}{RGB}{220,50,50}
\begin{table*}[t]
    \centering
    % \hspace{2mm}
    
    \setlength{\tabcolsep}{2.8pt}
    \renewcommand{\arraystretch}{1.4}
    \footnotesize
	\centering
    \begin{tabular}{p{2.6cm}|c c c c c c c c c| c}
        \shline
        \textbf{Method} & \textbf{GQA} & \textbf{MMB} & \textbf{MME} & \textbf{POPE} & \textbf{SQA} & \textbf{VQA}$^{\text{V2}}$ & \textbf{VQA}$^{\text{Text}}$ & \textbf{MMMU}& \textbf{SEED-I}   &\makecell[c]{\textbf{Avg}.}\\
        \shline
        \rowcolor{mygray}
        \multicolumn{11}{c}{\textit{Upper Bound, 576 Tokens} \ $\textbf{(100\%)}$}\\
        \multirow{2}*{Vanilla 7B} & 62.4 & 69.3 & 1841 &  85.8  & 70.7 & 80.4 & 65.2 & 36.1 & 69.7 & \multirow{2}*{100\%} \\
        ~ & 100\% & 100\% &100\% & 100\% & 100\% & 100\% & 100\% & 100\% & 100\%   & ~ \\
        \hline

        \rowcolor{mygray}
        \multicolumn{11}{c}{\textit{Retain 192 Tokens} \ $\fg{(\downarrow 66.7\%)}$} \\

        \multirow{2}*{VisionZip} & 60.3 & 68.9 & 1846 &82.3 & 70.1& 79.1 &63.4 & 36.1  &67.5 & \multirow{2}*{\textbf{98.2\%}} \\
        ~ & 96.6\% & 99.4\% & 100.2\% &95.9\% &99.2\% & 98.4\% & 97.2\% & 100\% & 96.8\%& ~\\
        \hline
        \multirow{2}*{VisionZip \ddag} & 61.6 & 67.2 & 1804 &85.5 & 70.2& 78.9 &63.6 & 36.1  &67.0 & \multirow{2}*{\textbf{98.3\%}} \\
        ~ & 98.7\% & 97.0\% &98.0\% &99.7\% &99.3\% & 98.1\% & 97.5\% & 100\% & 96.1\%& ~\\
        \hline
        \rowcolor{mygray}
        \multicolumn{11}{c}{\textit{Retain 128 Tokens} \ $\fg{(\downarrow 77.8\%)}$}\\
        
        \multirow{2}*{VisionZip} & 58.7 & 68.1 & 1841 &78.5&70.0 & 77.5& 61.3& 34.8 &65.6    & \multirow{2}*{\textbf{96.0\%}} \\
        ~ & 94.1\% & 98.3\% &100\% &91.5\% &99.0\% & 96.4\% & 94.0\% & 96.4\% & 94.1\%& ~\\
        \hline
        \multirow{2}*{VisionZip \ddag} & 60.0 & 67.0 & 1810 & 83.2 &70.1& 78.3 &61.6 & 34.8  &65.9 & \multirow{2}*{\textbf{96.7\%}} \\
        ~ & 96.2\% & 96.7\% &98.3\% &97.0\% &99.2\% & 97.4\% & 94.5\% & 96.4\% & 94.5\%& ~\\
        \hline
        \rowcolor{mygray}
        \multicolumn{11}{c}{\textit{Retain 64 Tokens} \ $\fg{(\downarrow 88.9\%)}$}\\
        
        \multirow{2}*{VisionZip} & 55.8 & 65.9 & 1737 &69.6 & 70.7& 73.9 &59.1 & 35.6  &61.7 & \multirow{2}*{\textbf{92.2\%}} \\

        ~ & 89.4\% & 95.1\% &94.4\% &81.4\% &100\% & 91.9\% & 90.6\% & 98.6\% & 88.5\%& ~\\
        \hline
        \multirow{2}*{VisionZip \ddag} & 57.7 & 66.3 &1779 & 80.0 &71.0 &  75.9 &60.1 & 36.2  &62.6 & \multirow{2}*{$\fr{95.0\%}$} \\
        ~ & 92.5\% & 95.7\% &96.6\% &93.2\% &100.4\% & 94.4\% & 92.2\% & 100.3\% & 89.8\%& ~\\

        \shline
	\end{tabular}
    % \vspace{-0.15cm}
	\caption{\textbf{Performance of ~\methodname\ on mini-Gemini 7B.} The vanilla number of visual tokens is $576$. The first line of each method shows the raw benchmark accuracy, and the second line is the proportion relative to the upper limit. The last column is the average value. \methodname\ddag~indicates that fine-tuning the multimodal projector with $1/10$ LLaVA-1.5 datasets. SEED-I represents SEED-IMG, which uses the metric from LMMs-Eval~\cite{zhang2024lmms}.}
	\label{tab:mgm-7b-full}
    % \vspace{-0.5cm}
\end{table*}

\subsubsection{Evaluation Benchmark}
\label{a:dataset}
We conducted experiments on these widely used visual understanding benchmarks.

\mypara{SEEDBench.}
SEEDBench~\cite{li2023seed} comprises 19,000 multiple-choice questions annotated by human assessors. The evaluation spans 12 distinct aspects, assessing the models’ ability to recognize patterns in images and videos across both spatial and temporal dimensions.

\mypara{MMMU.}
MMMU~\cite{yue2023mmmu} evaluates multimodal models on complex tasks requiring college-level knowledge and reasoning. It includes 11.5K curated questions from exams, quizzes, and textbooks, spanning six disciplines: Art \& Design, Business, Science, Health \& Medicine, Humanities \& Social Science, and Tech \& Engineering. Covering 30 subjects and 183 subfields, these questions incorporate 30 image types like charts, diagrams, and chemical structures. MMMU challenges models with advanced perception and domain-specific reasoning, similar to expert-level.

\mypara{MMVet.}
MMVet~\cite{yu2024mm} defines six core vision-and-language (VL) capabilities: recognition, OCR, knowledge, language generation, spatial awareness, and math. These capabilities integrate to address a range of complex multimodal tasks. MM-Vet evaluates 16 specific integrations of these capabilities through quantitative assessments.

\mypara{LLaVA-Bench.}
LLaVA-Bench~\cite{liu2023improvedllava} collects a diverse set of 24 images paired with 60 questions, encompassing indoor and outdoor scenes, memes, paintings, sketches, and more. Each image is accompanied by a highly detailed, manually curated description and a carefully selected set of questions. This design also evaluates the model's robustness to various prompts. Additionally, LLaVA-Bench categorizes questions into three types: conversational (simple QA), detailed description, and complex reasoning.

\mypara{VizWiz.}
VizWiz~\cite{gurari2018vizwiz} comprises over 31,000 visual questions created by blind individuals, each capturing a photo using a mobile phone and recording a spoken question about it. Each visual question is paired with 10 crowdsourced answers. The images, taken by blind photographers, are often of lower quality, the questions are spoken and conversational, and some visual questions cannot be answered due to the nature of the content.

\mypara{MMBench.} 
MMBench~\cite{liu2023mmbench} evaluates models through three hierarchical levels of abilities: L-1 with two core abilities (perception and reasoning), L-2 with six sub-abilities, and L-3 with 20 specific dimensions. This structure enables a detailed assessment of diverse capabilities.

\mypara{ScienceQA.}
Spanning domains like natural, language, and social sciences, ScienceQA~\cite{lu2022learn} organizes questions hierarchically into 26 topics, 127 categories, and 379 skills. This benchmark evaluates multimodal understanding, multi-step reasoning, and interpretability.

\mypara{GQA.}
The GQA~\cite{hudson2019gqa} benchmark evaluates visual scene understanding and reasoning using scene graphs, questions, and images. It includes spatial attributes and object features, with questions designed to test interpretation and reasoning.

\mypara{POPE.} 
POPE~\cite{li2023evaluating} evaluates Object Hallucination in models using binary questions on object presence in images. Metrics like Accuracy, Recall, Precision, and F1 Score measure hallucination levels across three sampling strategies, offering precise assessments.

\mypara{MME.}
The MME~\cite{fu2023mme} benchmark evaluates model performance across 14 subtasks targeting perceptual and cognitive abilities. Using manually designed instruction-answer pairs, MME minimizes data leakage for fair assessment.

\mypara{VQA-V2.} 
VQA-V2~\cite{goyal2017making} tests visual perception using 265,016 images of real-world scenes and objects paired with open-ended questions. Each question includes 10 ground truth answers from human annotators for accurate evaluation.

\mypara{TextVQA.} 
TextVQA~\cite{singh2019towards} evaluates a model's ability to interpret visual elements and embedded text in images through tasks requiring reasoning with textual information for accurate answers.

\subsubsection{Additional Experiments for LLaVA-1.5}
\mypara{Effectiveness on 13B. }
In the main paper, we demonstrate the effectiveness of our model on 7B in Table~\ref{tab:llava1-5}, and we show the effectiveness of our model on 13B in this section.
As shown in Table~\ref{tab:llava1-5-13b}, we conduct our proposed~\methodname~on 11 widely used evaluation benchmark. 
Due to the small size of LLaVA-Bench (LLaVA Wild Bench) and MMVeT, as well as the observation that their results can sometimes be unstable, we have excluded them from the average calculation in the last column. This decision was made despite our method demonstrating strong performance on both benchmarks. Instead, the average is calculated exclusively based on the 9 benchmarks.
As shown in Table~\ref{tab:llava1-5-13b}, we evaluate our method on three configurations of the vision token count (192, 128, and 64). The results show that even when retaining only 64 visual tokens, our method achieves 93.7\% performance without requiring additional training time. In the efficient-tuning mode, this performance increases to 94.8\%. Furthermore, when retaining 128 or 192 tokens, our method shows almost no performance loss in the 13B model.

\mypara{Effectiveness on Training Stage. }
Our proposed method can also be applied during the training stage to reduce token length, thereby saving memory usage and training time.
As shown in Table~\ref{tab:llava1-5-train-supp}, we conduct experiments on three different vision token count configurations (192, 128, and 64). We apply our proposed \methodname\ during the fine-tuning stage~\cite{liu2023improvedllava}, with all hyperparameters, except for the batch size, following the vanilla training settings. All experiments are conducted on 8 Nvidia 3090 24G GPUs with a batch size of 4.
To demonstrate the effectiveness of VisionZip in training mode, we evaluate it on 12 benchmarks and present the results. However, when calculating the average, we exclude LLaVA-Bench (LLaVA Wild Bench) and MMVet due to its small size and the observation that its results can be unstable, even though our method performs strongly on it. 
The results show that even when the number of tokens is reduced to 128, 99.6\% of the performance is retained. When retaining 192 tokens, performance even improves by 0.6\%.
We believe the reason is that reducing the redundancy of input visual tokens and providing only the more informative ones minimizes interference from less informative tokens. This allows the model to focus more on the informative tokens during training, enhancing visual understanding and leading to improved performance.

\subsubsection{Additional Experiments for LLaVA-NeXT}
In the main paper Table~\ref{tab:llava1-6}, we present the performance of ~\methodname\ on LLaVA-NeXT across several evaluation benchmarks. The complete benchmark results are provided in Table~\ref{tab:llava1-6-7b-full}. In this table, we only display the LLaVA NeXT 7B results for these stable benchmarks, and the results demonstrate that our proposed VisionZip consistently delivers strong performance.

To further demonstrate the effectiveness of our \methodname, we present the results on the LLaVA-NeXT 13B model.
As shown in Table~\ref{tab:llava1-6-13b-full}, our method demonstrates excellent scalability. As the size of the LLM increases, the performance of ~\methodname\ does not degrade. Our proposed VisionZip is highly adaptable to various sizes and types of LLMs, further highlighting the effectiveness of our approach. Notably, when retaining only 640 tokens, which eliminating 77.8\% of the tokens, our method enables the 13B model to outperform the 7B model in training-free mode. Furthermore, the generation speed of our 13B model is faster, and we will provide detailed speed in the next section.

\subsubsection{Additional Experiments for Mini-Gemini}
In the main paper, Fig.~\ref{fig:mgm} demonstrates that our method outperforms approaches like SparseVLM and FastV in terms of performance. Furthermore, as the number of retained tokens decreases, the performance advantage of our method becomes increasingly significant. In this section, we provide a detailed analysis of the results achieved by our method.

As shown in Table~\ref{tab:mgm-7b-full}, the results indicate that after removing 88.9\% of the tokens, our method can still retain over 90\% of its performance in the training-free mode. Furthermore, with fine-tuning, its performance can reach up to 95\%. When discarding 66.7\% of the visual tokens, which is more than half, the performance remains virtually unaffected. These results further highlight the significant redundancy present in visual tokens.

\subsubsection{Ablation Study}
\renewcommand{\multirowsetup}{\centering}
\definecolor{mygray}{gray}{.92}
\definecolor{ForestGreen}{RGB}{34,139,34}

\definecolor{Forestred}{RGB}{220,50,50}

\begin{table}[t]
    \centering
    \setlength{\tabcolsep}{1.5pt} 
    \renewcommand{\arraystretch}{1.2} 
    \small 
    \resizebox{\columnwidth}{!}{
        \begin{tabular}{
            @{} 
            l | 
            *{7}{c} | 
            c
            @{}
        }
            \toprule
            \textbf{Dataset} & \textbf{GQA} & \textbf{MMB} & \textbf{MME} & \textbf{SQA} & \textbf{VQA}$^{\text{V2}}$ & \textbf{VQA}$^{\text{Text}}$ & \textbf{MMMU} & \makecell[c]{\textbf{Avg}.} \\
            \hline
            \rowcolor{mygray}
            \multicolumn{9}{c}{\textit{Retain 640 Tokens} \ $\fg{(\downarrow 77.8\%)}$} \\

            LLaVA-1.5 & 62.4 & 65.9 & 1778 & 67.9 & 79.9 & 60.8 & 37.2 & 98.9\% \\
            LLaVA-NeXT &63.0 & 66.8 & 1738 & 68.4 & 80.1 & 61.2 & 38.8 & 99.3\%  \\
            \hline
            \rowcolor{mygray}
            \multicolumn{9}{c}{\textit{Retain 320 Tokens} \ $\fg{(\downarrow 88.9\%)}$} \\
            
            LLaVA-1.5  & 61.0 & 64.4 & 1770 & 67.5 & 78.4 & 59.3 & 38.0 & 97.6\% \\
            LLaVA-NeXT  & 61.6 & 64.7 & 1771 & 67.5 & 78.8 & 60.1 & 36.3 & 97.3\% \\             
            \hline
            \rowcolor{mygray}
            \multicolumn{9}{c}{\textit{Retain 160 Tokens} \ $\fg{(\downarrow 94.4\%)}$} \\
            LLaVA-1.5 & 58.2 & 63.9 & 1699 & 67.5 & 75.6 & 57.3 & 37.7 & 95.2\% \\
            LLaVA-NeXT & 58.4 & 63.2 & 1763 & 68.0 & 76.0 & 58.2 & 36.9 & 95.7\% \\  
            \bottomrule
        \end{tabular}}
        % \vspace{-0.3cm}
    \caption{\textbf{Impact of Fine-Tuning Dataset Compatibility.} The first column indicates which dataset was used to sample 1/10 of the data for fine-tuning the multimodality projector.}
    \label{tab:ablation}
% \vspace{-0.3cm}
\end{table}

\mypara{Impact of Fine-Tuning Dataset Compatibility}
We use \methodname\ to efficiently fine-tune the cross-modality projector, addressing the gap caused by reduced visual tokens. Ensuring dataset compatibility with the original model is crucial for optimal performance. To evaluate this, we compare the effects of using $1/10$ of the LLaVA 1.5 and LLaVA-NeXT datasets to fine-tune the LLaVA-NeXT model across three token count configurations (640, 320 and 160).
As shown in Table~\ref{tab:ablation}, improving dataset compatibility results in minimal gains (less than 0.5\%), with performance on some benchmarks even declining. These findings suggest that for efficient tuning to address token reduction, the basic $1/10$ LLaVA 1.5 dataset is sufficient.
The results further demonstrate that the performance gains of ~\methodname$\ddag$ in Table~\ref{tab:llava1-5} and Table~\ref{tab:llava1-6} of the main text are not attributable to additional knowledge acquired through continued training. Instead, these improvements arise from adaptation to the sudden reduction in tokens, which helps bridge the gap between the visual and LLM spaces. This finding aligns with our motivation outlined in Sec.~\ref{sec:readapt}.

% \mypara{Ablation Study for Dominant Tokens}

\subsection{Video Understanding}
\subsubsection{Evaluation Benchmark}
\mypara{TGIF-QA.} 
TGIF-QA~\cite{jang2017tgif} extends ImageQA to videos with 165,000 question-answer pairs based on GIFs. It includes three VideoQA tasks—repetition count, repeating action, and state transition—requiring spatio-temporal reasoning, plus frame QA tasks answerable from single frames.

\mypara{MSVD-QA.} 
MSVD-QA~\cite{xu2017video}, based on the MSVD dataset, features 1,970 video clips and 50.5K question-answer pairs. Covering diverse topics, it supports video question answering and captioning with open-ended questions in five categories: what, who, how, when, and where.

\mypara{MSRVTT-QA.} 
MSRVTT-QA~\cite{xu2017video} includes 10,000 video clips and 243,000 question-answer pairs, emphasizing video understanding and reasoning. Questions, categorized into what, who, how, when, and where, require models to process visual and temporal information.

\mypara{ActivityNet-QA.} 
ActivityNet-QA~\cite{yu2019activityqa} consists 58,000 human-annotated question-answer pairs from 5,800 ActivityNet videos. Covering motion, spatial, and temporal relationships, it evaluates VideoQA models on long-term spatio-temporal reasoning.

\begin{figure*}
    \centering
    \includegraphics[width=1\linewidth]{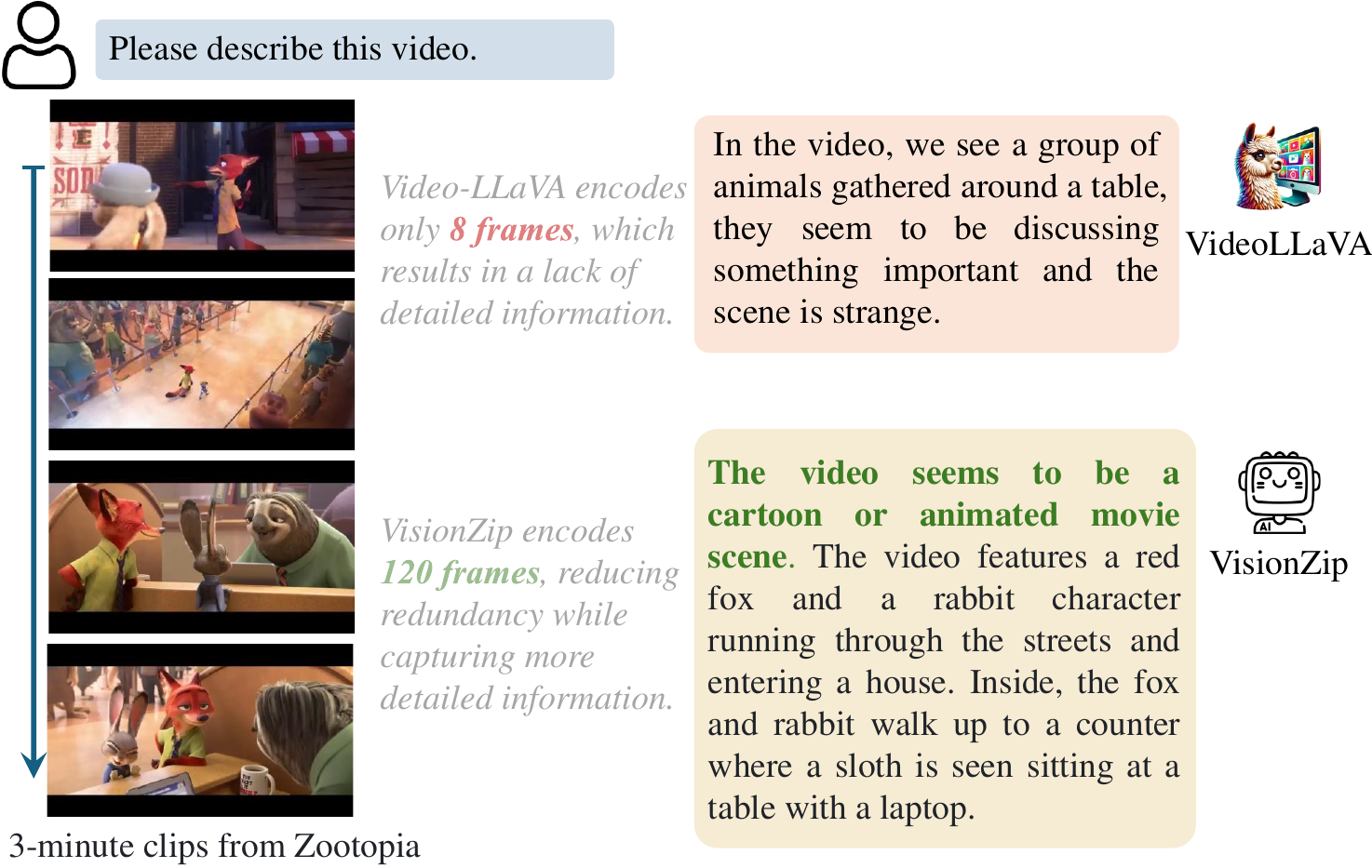}
    \caption{\textbf{Advantage of ~\methodname~in video understanding task.} With the same visual token length, using~\methodname~allows encoding more frames, significantly enhancing the model's capacity to understand longer video sequences and capture more detailed information. }
    \label{fig:future_supp}
\end{figure*}
\subsubsection{Future Direction}
With the development of LLMs and VLMs, video understanding has become a popular research direction. Whether the goal is for VLMs to comprehend longer videos or to achieve precise localization within videos, enabling the input of more frames within limited memory is both important and critical.

However, existing methods process a single frame into at least 256 tokens, which hinders the ability to input more frames. With our approach, VisionZip, the number of video tokens can be reduced by 5-10 times before being input into the LLM. This reduction allows the model to process 5-10 times more frames within the same memory constraints. For example, if a model could originally handle only 1 hour of video, VisionZip enables it to process 5-10 hours of video, significantly enhancing the application value of VLMs in video understanding.

As shown in Fig.~\ref{fig:future_supp}, we select 3-minute video clips from Zootopia, a well-known cartoon, and ask the model to describe it. The results show that VideoLLaVA tends to describe a single frame in detail, lacking an overall understanding of the video, as it can only encode an 8-frame video. In contrast, our \methodname\ can encode 10$\times$ more video frames without increasing the token count, significantly enhancing the model's ability to understand longer videos.

\renewcommand{\multirowsetup}{\centering}
\definecolor{mygray}{gray}{.92}
\definecolor{ForestGreen}{RGB}{34,139,34}
\definecolor{Forestred}{RGB}{220,50,50}
\begin{table*}[t]
    \centering
    % \hspace{2mm}
    
    \setlength{\tabcolsep}{2.8pt}
    \renewcommand{\arraystretch}{1.4}
    \footnotesize
	\centering
    \begin{tabular}{p{2cm}|c |c c c c c c c c c| c}
        \shline
        \textbf{Method} & \textbf{Memory} &  \textbf{GQA} & \textbf{MMB} & \textbf{MME} & \textbf{POPE} & \textbf{SQA} & \textbf{VQA}$^{\text{V2}}$ & \textbf{VQA}$^{\text{Text}}$ & \textbf{MMMU}& \textbf{SEED-I}   &\makecell[c]{\textbf{Avg}.}\\
        \shline
        \rowcolor{mygray}
        \multicolumn{12}{c}{\textit{Upper Bound, 2880 Tokens} \ $\textbf{(100\%)}$}\\
        \multirow{2}*{Vanilla 13B} &\multirow{2}*{36721Mb} & 65.4& 70.0 & 1901 &  86.2  & 73.5 & 81.8 & 64.3 & 36.2 & 71.9 & \multirow{2}*{100\%} \\
        ~ & ~ &100\% & 100\% & 100\% & 100\% & 100\% & 100\% & 100\% & 100\% & 100\%   & ~ \\
        \hline
        \multirow{2}*{Vanilla 7B} & \multirow{2}*{18952Mb} & 64.2 & 67.9 & 1842 &  86.4  & 70.2 & 80.1 & 61.3 & 35.1 & 70.2 & \multirow{2}*{97.2\%} \\
        ~ & ~ &98.2\% & 96.3\% & 96.9\% & 100.2\% & 95.5\% & 97.9\% & 95.3\% & 97.0\% & 97.6\%   & ~ \\
        \hline

        \rowcolor{mygray}
        \multicolumn{12}{c}{\textit{Retain 320 Tokens} \ $\fg{(\downarrow 88.9\%)}$}\\
        
        \multirow{2}*{VisionZip} & \multirow{2}*{28810Mb} &60.7 & 67.2 &1805 & 82.0 &70.3& 76.8 &60.9 & 35.6& 65.2& \multirow{2}*{\textbf{94.7\%}} \\
        ~ & ~ &92.8\% & 96.0\% & 95.0\% &95.1\% & 95.6\% & 93.9\% & 94.7\% & 98.3\% & 90.7\% & ~\\
        \hline
        \multirow{2}*{VisionZip-8bit} & \multirow{2}*{16632Mb} & 60.6 &67.1 & 1798  &81.4& 70.8 &76.8 & 60.5 & 37.0 & 65.4& \multirow{2}*{\textbf{95.0\%}} \\
        ~ & ~ &92.7\% & 95.9\% & 94.6\% & 94.4\% & 96.3\% & 93.9\% & 94.1\% & 102.2\% & 91.0\% & ~ \\
        \hline
        \multirow{2}*{VisionZip-4bit} & \multirow{2}*{10176Mb} & 60.3 &65.1 & 1773  &82.1& 70.3 &76.6  &60.0& 36.1 & 65.1& \multirow{2}*{\textbf{94.0\%}} \\
        ~ & ~ &92.2\% & 93.0\% & 93.3\% &95.2\% & 95.6\% & 93.6\% & 93.3\% & 99.7\% & 90.5\% & ~ \\
        
        \shline
	\end{tabular}
    % \vspace{-0.15cm}
	\caption{\textbf{Performance and Memory of ~\methodname\ on LLaVA NeXT 13B with the Quantization.} The vanilla number of visual tokens is $2880$. The first line of each method shows the raw benchmark accuracy, and the second line is the proportion relative to the upper limit. The last column is the average value. SEED-I represents SEED-IMG, which uses the metric from LMMs-Eval~\cite{zhang2024lmms}. The memory refers to the practical CUDA memory usage on a single Nvidia A800 GPU for SQA.}
	\label{tab:llava1-6-13b-bit-supp}
    % \vspace{-0.5cm}
\end{table*}

\renewcommand{\multirowsetup}{\centering}
\definecolor{mygray}{gray}{.92}
\definecolor{ForestGreen}{RGB}{34,139,34}
\definecolor{Forestred}{RGB}{220,50,50}
\begin{table*}[t]
    \centering
    % \hspace{2mm}
    
    \setlength{\tabcolsep}{2.8pt}
    \renewcommand{\arraystretch}{1.4}
    \footnotesize
	\centering
    \begin{tabular}{c|c c | c c c c c c c c c| c}
        \shline
        \textbf{Method} & \textbf{Time}& \textbf{Memory} &  \textbf{GQA} & \textbf{MMB} & \textbf{MME} & \textbf{POPE} & \textbf{SQA} & \textbf{VQA}$^{\text{V2}}$ & \textbf{VQA}$^{\text{Text}}$ & \textbf{MMMU}& \textbf{SEED-I}   &\makecell[c]{\textbf{Avg}.}\\
        \shline
        \rowcolor{mygray}
        \multicolumn{13}{c}{\textit{Upper Bound, 2880 Tokens} \ $\textbf{(100\%)}$}\\
        \multirow{2}*{Vanilla 7B} & \multirow{2}*{33.8h} & \multirow{2}*{63558Mb} & 64.2 & 67.9 & 1842 &  86.4  & 70.2 & 80.1 & 61.3 & 35.1 & 70.2 & \multirow{2}*{100\%} \\
        ~ & ~ &~ &100\% & 100\% & 100\% & 100\% & 100\% & 100\% & 100\% & 100\% & 100\%   & ~ \\
        \hline

        \rowcolor{mygray}
        \multicolumn{13}{c}{\textit{Retain 640 Tokens} \ $\fg{(\downarrow 77.8\%)}$}\\
        
        \multirow{2}*{VisionZip-Inference} & \multirow{2}*{~} & \multirow{2}*{~} &61.3 & 66.3 &1787 & 86.3 &68.1& 79.1 &60.2 & 34.7& 66.7& \multirow{2}*{\textbf{97.5\%}} \\
        ~ & ~ ~ & &95.5\% & 97.6\% & 97.0\% &99.9\% & 97.0\% & 98.8\% & 98.2\% & 98.9\% & 95.0\%  & ~\\
        \hline
        \multirow{2}*{VisionZip-Train} & \multirow{2}*{15.9h} & \multirow{2}*{35326Mb}  & 62.5 &67.1 & 1728  &86.0& 70.2 &80.6 & 64.1 & 35.1 & 67.8& \multirow{2}*{$\fr{99.0\%}$} \\
        ~ & ~ & ~ &97.4\% & 98.8\% & 93.8\% & 99.5\% & 100\% & 100.6\% & 104.6\% & 100\% & 96.6\% & ~ \\
        
        \shline
	\end{tabular}
    % \vspace{-0.15cm}
	\caption{\textbf{Performance and Training Time of ~\methodname\ on LLaVA NeXT 7B.} The vanilla number of visual tokens is $2880$. The first line of each method shows the raw benchmark accuracy, and the second line is the proportion relative to the upper limit. The last column is the average value. SEED-I represents SEED-IMG, which uses the metric from LMMs-Eval~\cite{zhang2024lmms}. The time refers to the practical Training time usage on  8 Nvidia A800 GPUs for training.}
	\label{tab:llava1-6-7b-train-supp}
    % \vspace{-0.5cm}
\end{table*}

\renewcommand{\multirowsetup}{\centering}
\definecolor{mygray}{gray}{.92}
\definecolor{ForestGreen}{RGB}{34,139,34}
\definecolor{Forestred}{RGB}{220,50,50}
\begin{table*}[t]
    \centering
    % \hspace{2mm}
    
    \setlength{\tabcolsep}{2.8pt}
    \renewcommand{\arraystretch}{1.4}
    \footnotesize
	\centering
    \begin{tabular}{c|ccc|c c c c c c c c c| c}
        \shline
        \textbf{Method} & Count& Prefilling & Total  &\textbf{GQA} & \textbf{MMB} & \textbf{MME} & \textbf{POPE} & \textbf{SQA} & \textbf{VQA}$^{\text{V2}}$ & \textbf{VQA}$^{\text{Text}}$ & \textbf{MMMU}& \textbf{SEED-I}   &\makecell[c]{\textbf{Avg}.}\\
        \shline

        \multirow{2}*{Vanilla 13B} &\multirow{2}*{2880}&\multirow{2}*{129.4ms} & \multirow{2}*{2506s}&65.4& 70.0 & 1901 &  86.2  & 73.5 & 81.8 & 64.3 & 36.2 & 71.9 & \multirow{2}*{100\%} \\
        ~ &~ &~ &~ & 100\% & 100\% & 100\% & 100\% & 100\% & 100\% & 100\% & 100\% & 100\%   & ~ \\
        \hline
        \multirow{2}*{Vanilla 7B} &\multirow{2}*{2880}&\multirow{2}*{54.2ms} &\multirow{2}*{1598s} & 64.2 & 67.9 & 1842 &  86.4  & 70.2 & 80.1 & 61.3 & 35.1 & 70.2 & \multirow{2}*{97.2\%} \\
        ~ &~ &~ &~ & 98.2\% & 96.3\% & 96.9\% & 100.2\% & 95.5\% & 97.9\% & 95.3\% & 97.0\% & 97.6\%   & ~ \\
        \hline

        \multirow{2}*{\methodname\ 13B} &\multirow{2}*{640} &\multirow{2}*{48.2ms}& \multirow{2}*{1219s}& 63.0 & 68.6 & 1871 &85.7 & 71.2& 79.7 &62.2 & 36.4  &68.8 & \multirow{2}*{97.5\%} \\
        ~ &~ &~ &~&96.3\% & 98.0\% & 98.4\% &99.4\% &96.7\% & 96.9\% & 96.7\% & 100.5\% & 95.7\%& ~\\

        \hline
        \multirow{2}*{\methodname~13B}&\multirow{2}*{320}& \multirow{2}*{30.3ms}& \multirow{2}*{995s}& 60.7 & 67.2 &1805 & 82.0 &70.3& 76.8 &60.9 & 35.6& 65.2& \multirow{2}*{94.7\%} \\
        ~ &~ &~ &~&92.8\% & 96.0\% & 95.0\% &95.1\% & 95.6\% & 93.9\% & 94.7\% & 98.3\% & 90.7\% & ~\\

        \hline
        \multirow{2}*{\methodname~13B}&\multirow{2}*{160}& \multirow{2}*{23.9ms}&\multirow{2}*{888s} & 57.8 & 64.9 & 1739 & 76.6 &69.3& 72.4 &58.4 & 37.0 & 61.1& \multirow{2}*{91.3\%} \\
        ~ &~ &~ &~& 88.4\% & 92.7\% & 91.5\% & 88.9\% & 94.3\% & 88.5\% & 90.8\% & 102.2\% & 84.8\% & ~ \\

        \shline
	\end{tabular}
    % \vspace{-0.15cm}
	\caption{\textbf{Performance of ~\methodname\ on LLaVA NeXT 13B.} The vanilla number of visual tokens is $2880$. The first line of each method shows the raw benchmark accuracy, and the second line is the proportion relative to the upper limit. The last column is the average value. ``Prefilling" represents the prefilling time, and ``Total" represents the actual testing time of the model on the TextVQA benchmark.}
	\label{tab:llava1-6-13b-time}
    % \vspace{-0.5cm}
\end{table*}

\subsection{Efficiency Analysis}
\label{sec:supp-efficiency}
In this section, we provide additional results highlighting the efficiency gains brought by ~\methodname.

\mypara{CUDA Memory Save.}
We conduct experiments on the LLaVA-NeXT 13B model, retaining only 320 visual tokens. Additionally, to better illustrate the memory consumption changes introduced by this process, we simultaneously present the performance variations alongside the CUDA memory changes in ScienceQA.
The result aligns with Table~\ref{tab:advantage-quant} in the main paper.
As shown in Table~\ref{tab:llava1-6-13b-bit-supp}, the third row demonstrates that using \methodname\ can reduce CUDA memory consumption by more than 20\%. Additionally, employing 8-bit and 4-bit quantization further reduces memory usage. Moreover, our method integrates seamlessly with quantization techniques, and the performance of the quantized model is comparable to the original results.

\mypara{Training Time Save.}
Our proposed \methodname\ can also reduce training time. We conducted an experiment on LLaVA-NeXT 7B, retaining 640 visual tokens. As shown in Table~\ref{tab:llava1-6-7b-train-supp}, using \methodname\ during the training stage significantly reduces training time by $2\times$ and achieves better performance compared to applying \methodname\ only during the inference stage.

\mypara{Inference Time Save.}
To demonstrate the relationship between the number of remaining tokens and inference time, we conduct experiments on LLaVA-NeXT 13B. We configured three vision token counts: 640, 320, and 160, respectively.
We recorded the prefilling time and the actual testing time on the benchmark. Specifically, we use the TextVQA dataset to conduct the time measurements.
As shown in Table~\ref{tab:llava1-6-13b-time}, by using VisionZip to retain 640 tokens, the 13B model achieves faster inference than the 7B model while maintaining superior performance.

\section{Related Work}
\label{sec:supp-related}
\mypara{Vision-Language Models. }
Building on the success of LLMs~\cite{achiam2023gpt, touvron2023llama, bai2023qwen,chiang2023vicuna,longlora,li2024quickllamaqueryawareinferenceacceleration,zheng2024dapedataadaptivepositionalencoding,rafailov2024direct,lai2024stepdpo}, VLMs have made significant advancements~\cite{liu2023improvedllava, liu2024llavanext, li2024mini, tong2024cambrian, wang2023cogvlm,liu2024visual,lai2023lisa,zhang2024prompt,yang2023improved,regionblip,huang2024ffaa}. Popular VLM models, such as LLaVA~\cite{liu2023improvedllava} and mini-Gemini~\cite{li2024mini}, process visual tokens through a projector before inputting them into the LLM as a sequence.
However, real-world images are typically high-resolution and require a large number of tokens.
For example, LLaVA-NeXT processes $672 \times 672$ images into more than 2,000 tokens \cite{liu2024llavanext}. 
Moreover, handling videos or multiple images significantly increases token requirements~\cite{lin2023video,Maaz2023VideoChatGPT, li2024llamavid, huang2025lita,tang2023video,song2024moviechat}.
. Hence, it’s essential to discuss more efficient ways to extract information from visual tokens, rather than merely increasing their length. 

\mypara{Efficient Large Language Models. }
In the field of large language models (LLMs), various strategies have been developed to reduce tokens, thereby accelerating inference and optimizing key-value (KV) cache compression~\cite{han2023lm}. For example, StreamingLLM~\cite{xiao2023streamingllm} decreases the KV cache size by retaining only the attention sinks and the most recent tokens. FastGen~\cite{ge2023model} introduces an adaptive method for managing the KV cache, dynamically optimizing memory usage by adjusting retention strategies based on the behavior of attention heads. Similarly, the Heavy-Hitter Oracle (H2O)~\cite{zhang2023h2o} employs a scoring mechanism based on cumulative attention to selectively prune key-value pairs during the generation process. These methods aim to reduce token redundancy and enhance the efficiency of inference operations in LLMs.

\mypara{Efficient Vision Language Models. }
Recently, some studies~\cite{chen2024image, zhang2024sparsevlm, xing2024pyramiddrop, wen2024efficient, shi2023upop, he2024zipvl} have also recognized the redundancy in visual tokens and proposed various methods to address it. And most of these works identify redundancy based on the relatively low attention that LLM text tokens assign to visual tokens. Furthermore, these studies primarily achieve token reduction or KV cache compression by leveraging attention mechanisms between text and visual tokens during the LLM forward process. In contrast to these works, we find that the visual tokens generated by popular vision encoders exhibit significant redundancy. Our approach removes this redundancy before the tokens are input into the LLM. Additionally, in Sec.~\ref{sec:analysis} of the main paper, we provide a thorough comparison and analysis of our method against these text-relevant approaches.

\section{Visualization}
\label{sec:sup-visual}
\subsection{Visualization of Redundancy}
\label{sec:visual-redundancy}
\begin{figure*}
    \centering
    \includegraphics[width=0.8\linewidth]{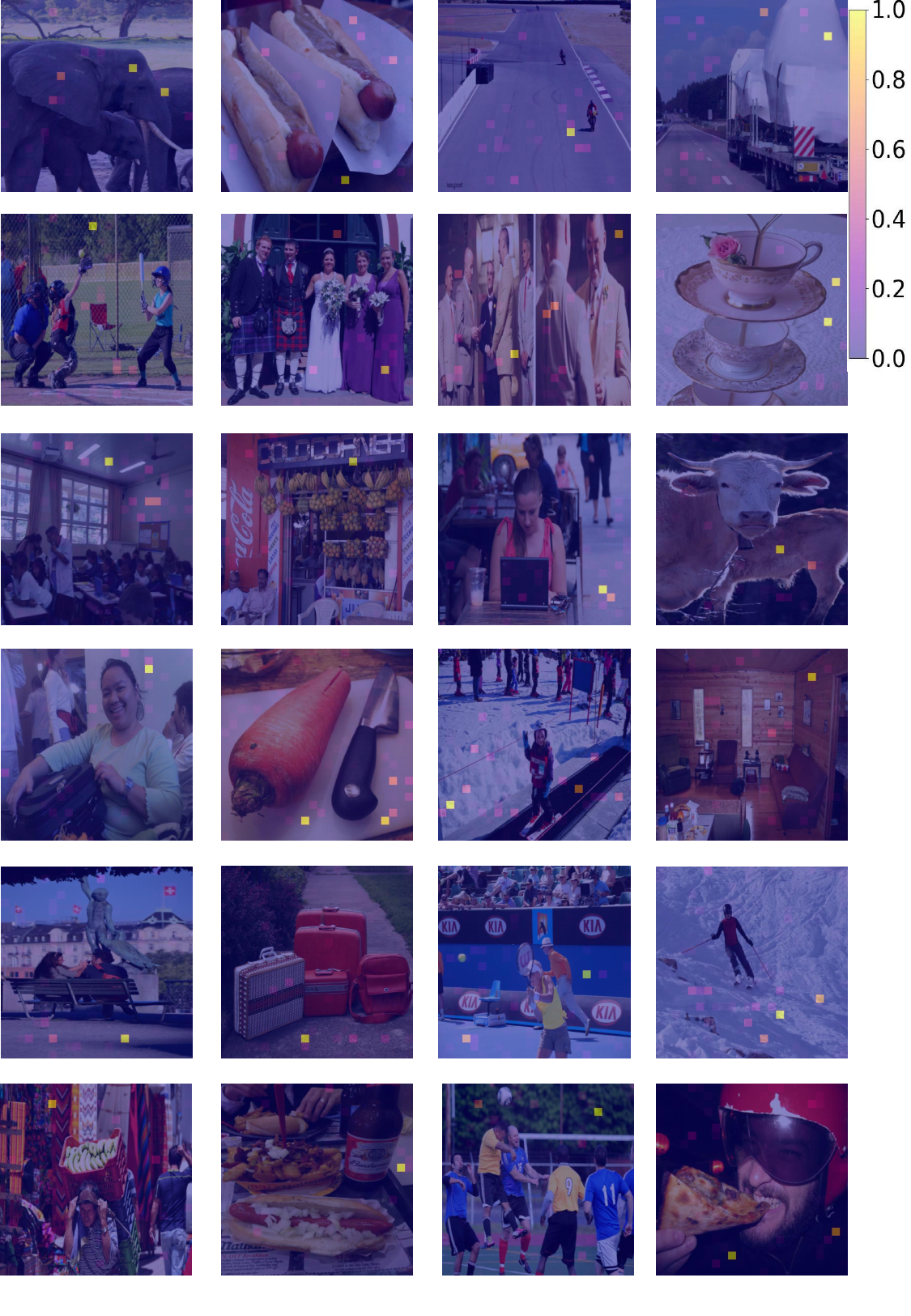}
    \caption{Visualization of Redundancy in the CLIP Model}
    \label{fig:Redundancy1}
\end{figure*}

\begin{figure*}
    \centering
    \includegraphics[width=0.8\linewidth]{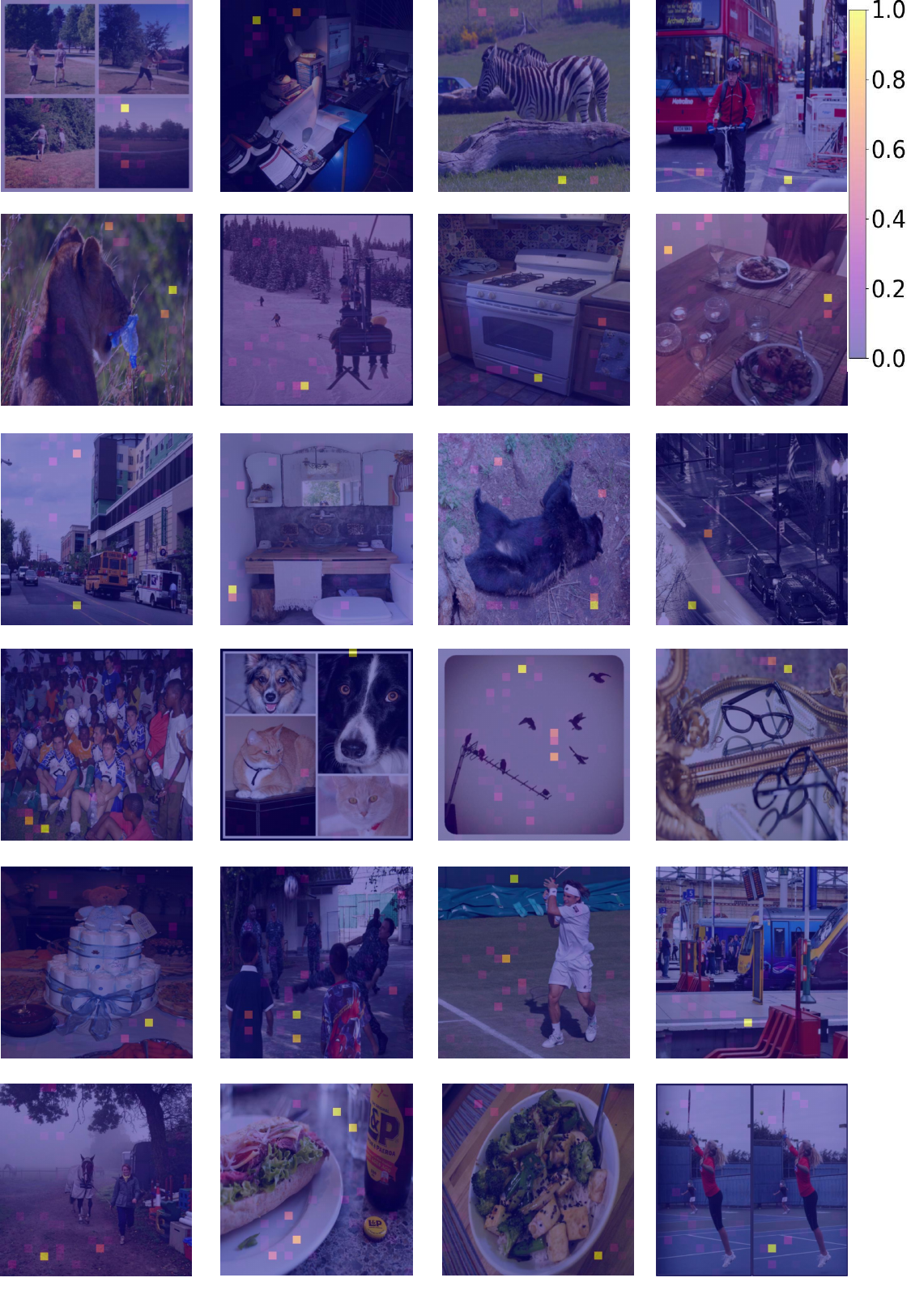}
    \caption{Visualization of Redundancy in the CLIP Model}
    \label{fig:Redundancy2}
\end{figure*}

\begin{figure*}
    \centering
    \includegraphics[width=0.8\linewidth]{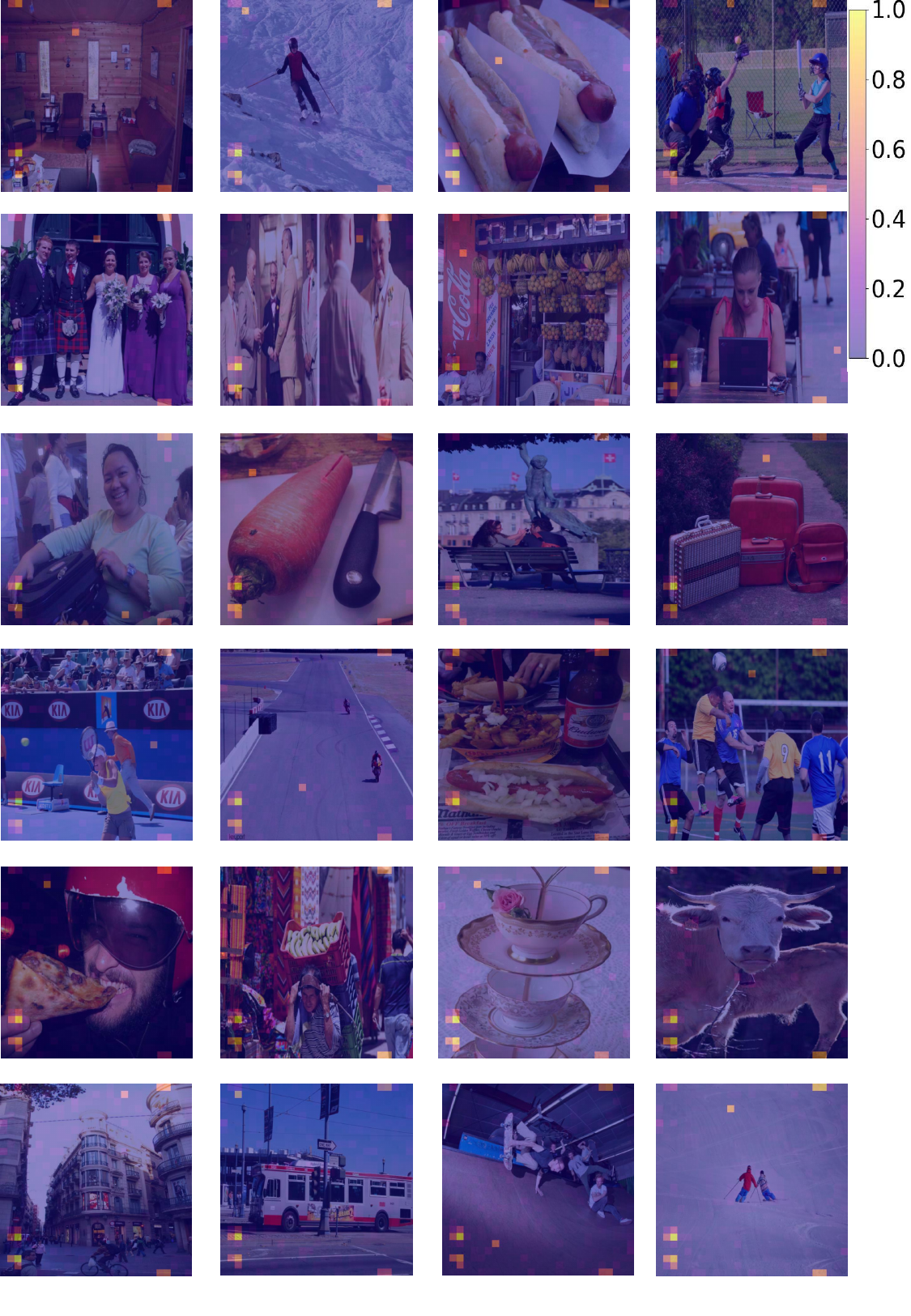}
    \caption{Visualization of Redundancy in the SigLIP Model}
    \label{fig:Redundancy1-siglip}
\end{figure*}

To further show the redundancy in popular vision encoders, we include additional examples from the COCO train2017 dataset. This dataset is a key component of the LLaVA 1.5 fine-tuning dataset and an essential part of many vision datasets.
As shown in Fig.~\ref{fig:Redundancy1}  Fig.~\ref{fig:Redundancy2} and Fig.~\ref{fig:Redundancy1-siglip},  the visualization results indicate that only a few tokens receive high attention and contain substantial amounts of information, while most visual tokens receive minimal attention and contain limited information. This visualization highlights the significant redundancy present in the visual tokens.

\subsection{Visualization of Attention Distribution Change}
\begin{figure*}
    \centering
    \includegraphics[width=0.8\linewidth]{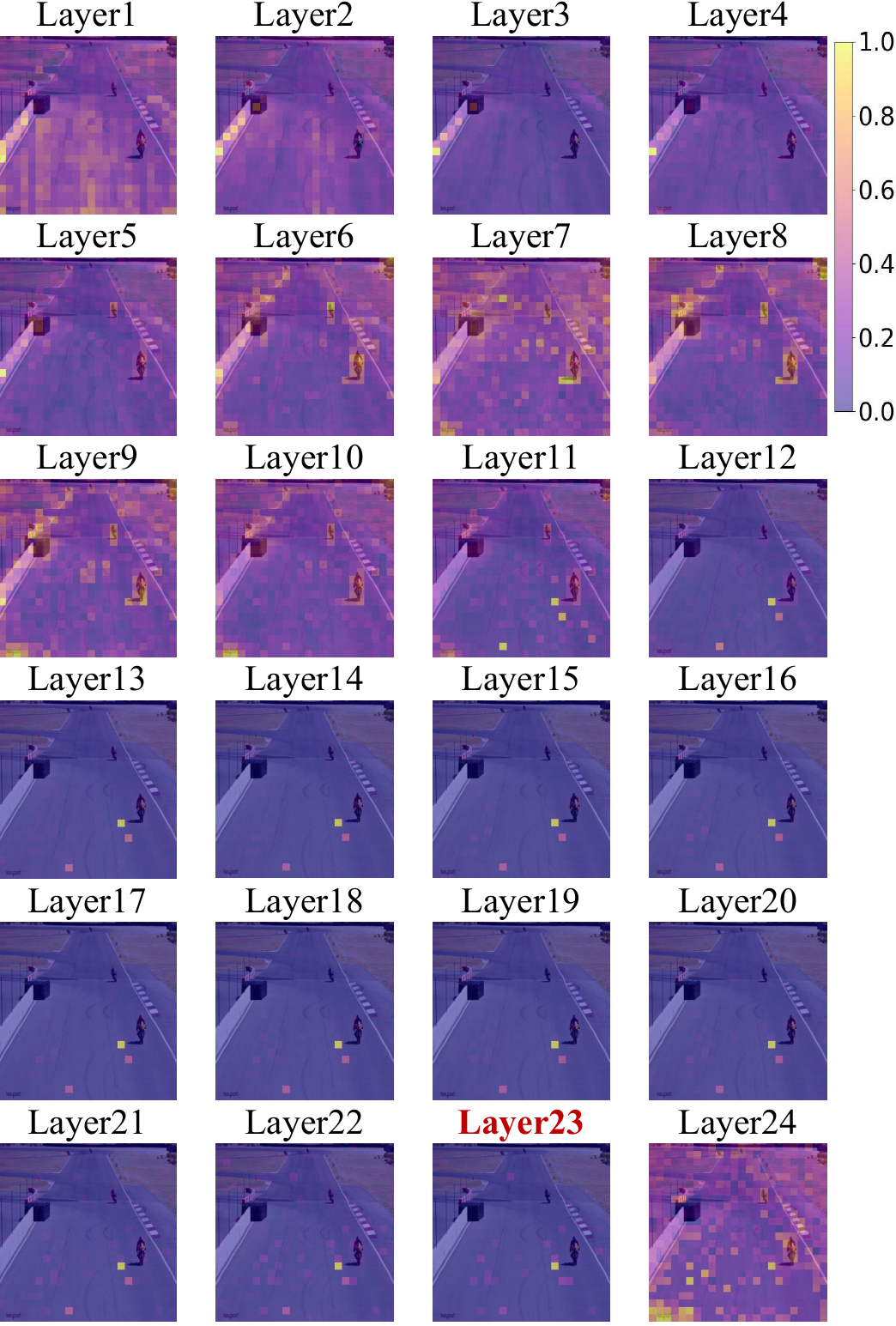}
    \caption{Visualization of Attention Distribution Change}
    \label{fig:distribution-Supp-1}
\end{figure*}

\begin{figure*}
    \centering
    \includegraphics[width=0.8\linewidth]{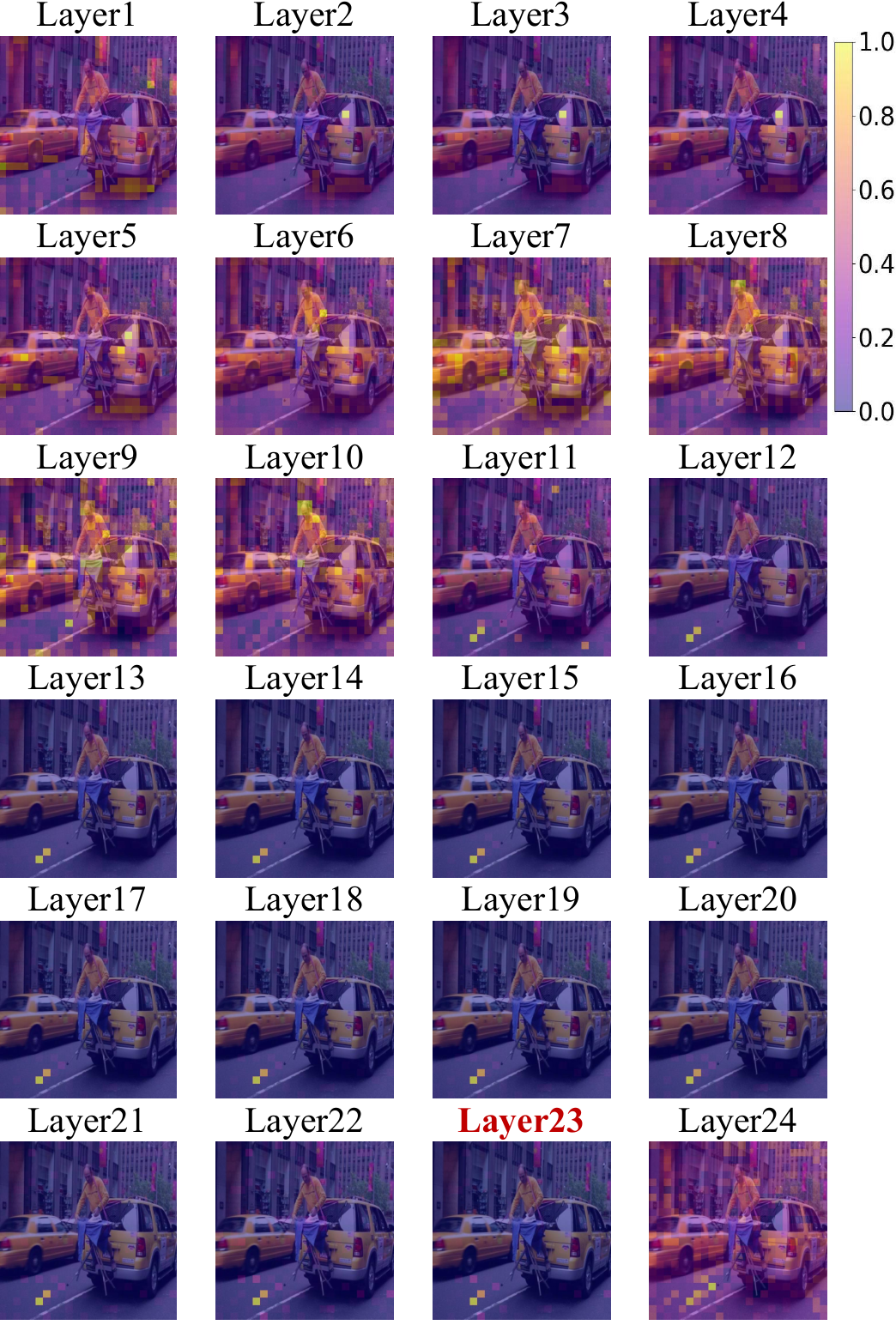}
    \caption{Visualization of Attention Distribution Change}
    \label{fig:distribution-Supp-2}
\end{figure*}

In Sec.~\ref{sec:analysis} of the main text, we discuss the reasons behind the redundancy in visual tokens. In this section, we present a comprehensive analysis of the changes in attention within the CLIP model.
As shown in Fig.~\ref{fig:distribution-Supp-1} and Fig.~\ref{fig:distribution-Supp-2} attention in the early layers is broadly distributed across the image. However, by the middle layers, it rapidly converges onto a few tokens. In the deeper layers, attention and information become concentrated on a small set of dominant tokens, reaching peak concentration by the 23rd layer, which is used for visual token extraction for the LLM. Besides, in the final layer, attention is more dispersed as these tokens align with the CLIP text branch via contrastive loss, potentially limiting their ability to represent the original image.

\subsection{Visualization of Feature Misalignment}
In Fig.~\ref{fig:softmax-misalign} of the main text, we show the phenomenon of feature misalignment. To further demonstrate that this phenomenon is widespread, we observe it across additional COCO images.

\begin{figure*}
    \centering
    \includegraphics[width=0.83\linewidth]{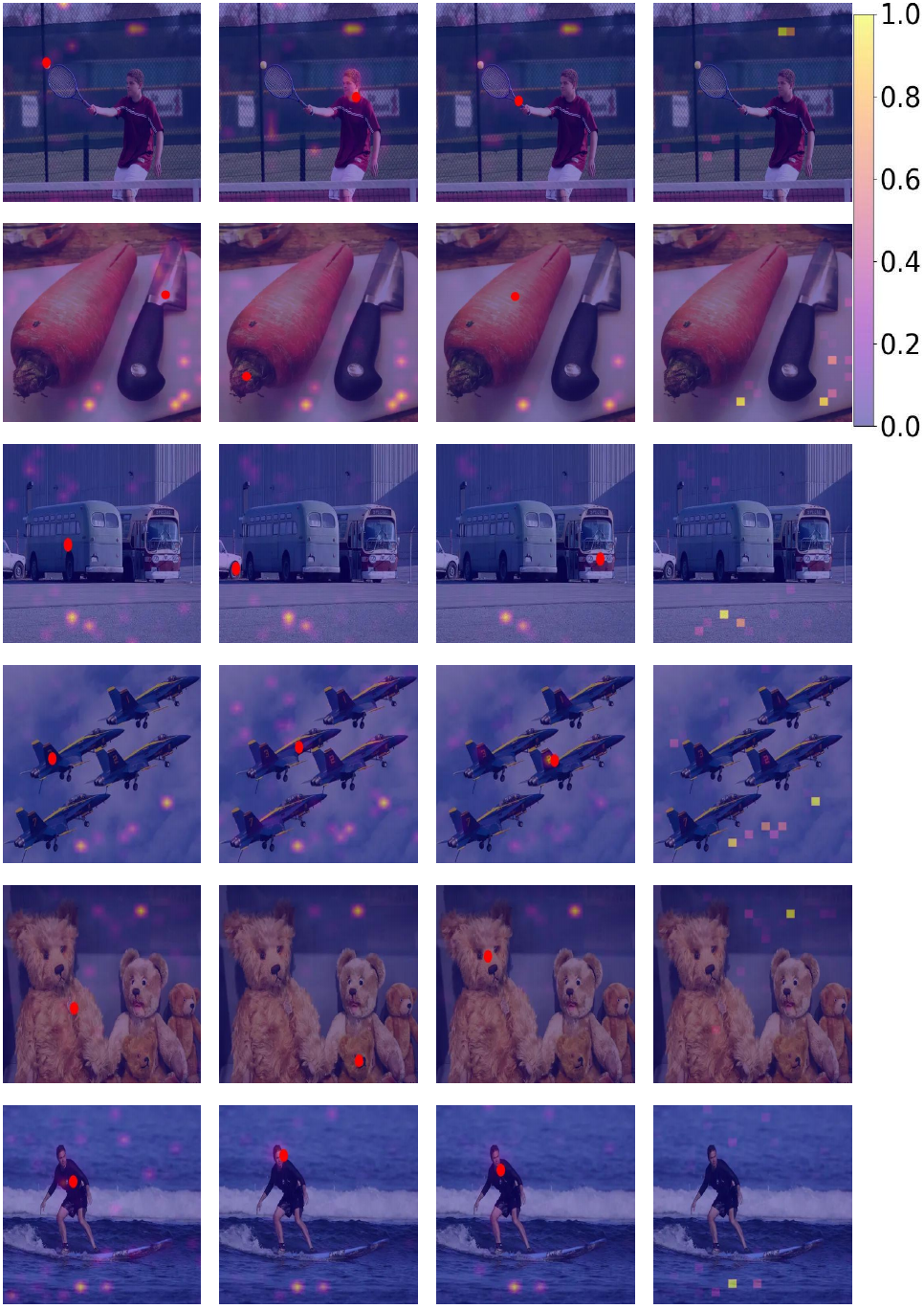}
    \caption{\textbf{Visualization of Feature Misalignment.} The red point represents the selected token, while the heatmaps in the first three columns illustrate the attention relationships to the selected token. The last column displays the attention map for the entire image. The results shows that the attention of the selected token does not focus on semantically similar tokens but instead on dominant tokens, highlighting the phenomenon of feature misalignment.}
    \label{fig:supp_misalign}
\end{figure*}

As shown in Fig.~\ref{fig:supp_misalign}, in the first three columns, we select a token (red point) from the main subject of the figure and illustrate the attention to that token, and the last column shows that the attention score for the whole figure. The results show that the attention of the selected token does not focus on semantically relevant tokens but instead on dominant tokens, highlighting the phenomenon of feature misalignment.
Hence, when text-relevant methods like SparseVLM~\cite{zhang2024sparsevlm} select tokens based on semantic relationships, they can identify semantically relevant tokens. However, these tokens contain less information compared to the dominant tokens, which aggregate information from the entire image.

\begin{figure*}
    \centering
    \includegraphics[width=1\linewidth]{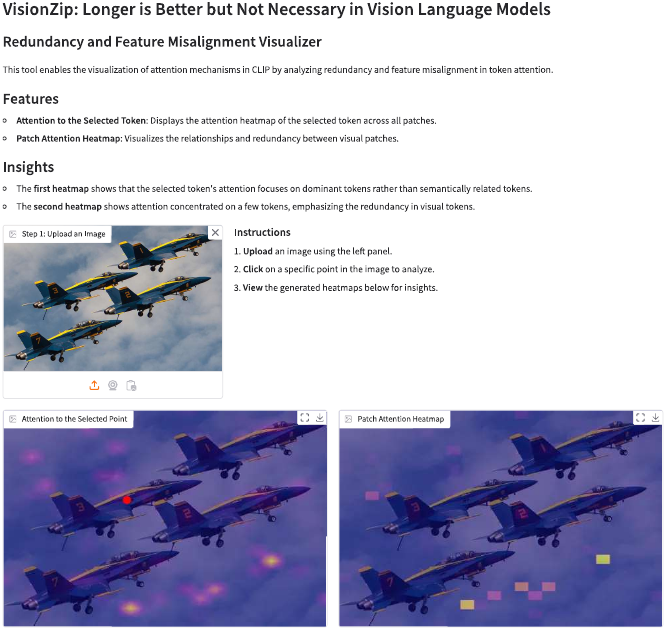}
    \caption{Gradio demo to analysis the visual redundancy and the feature misalignment}
    \label{fig:gradio-demo}
\end{figure*}

In addition, to improve visualization and analysis, we developed a Gradio demo, as shown in Fig.~\ref{fig:gradio-demo}. The corresponding code is provided on the GitHub page.


\end{document}